\documentclass{article}

\PassOptionsToPackage{numbers, compress}{natbib}

\usepackage[final]{neurips_2025}

\usepackage[utf8]{inputenc} 
\usepackage[T1]{fontenc}    
\usepackage{hyperref}       
\usepackage{url}            
\usepackage{booktabs}       
\usepackage{amsfonts}       
\usepackage{nicefrac}       
\usepackage{microtype}      
\usepackage[bottom]{footmisc}
\usepackage{graphicx}
\usepackage{multirow}
\usepackage{amsfonts}
\usepackage{adjustbox}
\usepackage[dvipsnames]{xcolor}

\usepackage{listings}
\usepackage{algorithm}
\usepackage{algpseudocode}

\usepackage{wrapfig}
\usepackage{caption}
\usepackage{subcaption,lipsum}
\usepackage{pifont}
\usepackage{bbm}
\usepackage{multicol}
\usepackage{afterpage}
\usepackage{placeins}
\usepackage{dblfloatfix}
\usepackage[flushleft]{threeparttable} %
\usepackage{enumitem}

\usepackage{amsmath}
\usepackage{amssymb}
\usepackage{mathtools}
\usepackage{amsthm}
\theoremstyle{plain}
\newtheorem{theorem}{Theorem}[section]
\newtheorem{proposition}[theorem]{Proposition}

\theoremstyle{definition}

\theoremstyle{remark}

\DeclarePairedDelimiterX{\infdivx}[2]{(}{)}{%
  #1\;\delimsize\|\;#2%
}

\title{Learning Without Augmenting: Unsupervised Time Series Representation Learning via Frame Projections}

\author{%
  Berken Utku Demirel \\
  Department of Computer Science \\
  ETH Zürich, Switzerland \\
  \texttt{berken.demirel@inf.ethz.ch} \\
  \And
  Christian Holz \\
  Department of Computer Science \\
  ETH Zürich, Switzerland\\
  \texttt{christian.holz@inf.ethz.ch} \\
}

\begin{document}

\maketitle

\begin{abstract}
Self-supervised learning (SSL) has emerged as a powerful paradigm for learning representations without labeled data. 
Most SSL approaches rely on strong, well-established, handcrafted data augmentations to generate diverse views for representation learning.
However, designing such augmentations requires domain-specific knowledge and implicitly imposes representational invariances on the model, which can limit generalization.
In this work, we propose an unsupervised representation learning method that replaces augmentations by generating views using orthonormal bases and overcomplete frames.
We show that embeddings learned from orthonormal and overcomplete spaces reside on distinct manifolds, shaped by the geometric biases introduced by representing samples in different spaces.
By jointly leveraging the complementary geometry of these distinct manifolds, our approach achieves superior performance without artificially increasing data diversity through strong augmentations.
We demonstrate the effectiveness of our method on nine datasets across five temporal sequence tasks, where signal-specific characteristics make data augmentations particularly challenging. 
Without relying on augmentation-induced diversity, our method achieves performance gains of up to 15--20\% over existing self-supervised approaches.
Source code: \url{https://github.com/eth-siplab/Learning-with-FrameProjections}
\end{abstract}

\section{Introduction}
\label{sec:intro}
Sample efficient unsupervised representation learning is a critical open challenge in deep learning.
While recent self-supervised techniques have shown strong performance in several tasks, they typically rely on handcrafted, aggressive data augmentations that expose the model to different versions of input samples in each training epoch to increase data diversity~\cite{simclr,BYOL,vicreg,simsiam}.
Moreover, these techniques only perform well when augmentations are carefully optimized for the downstream task; otherwise, model performance drops significantly~\cite{simclr, demirel2023chaos}.
Even when the downstream task is known, designing effective augmentations can be challenging for data types lacking a well-defined structure, i.e., text, tabular, signals~\cite{DACL, ssmix, sui2024selfsupervised, unsupervised_periodicity}, as the overly strong augmentations can cause model collapse~\cite{understand_collapse}.

Simply increasing sample diversity through random augmentations does not guarantee improved performance in representation learning.
Augmentations are only effective if the augmented views of a sample have sufficient representational similarity with views of other intra-class samples~\cite{X_lecun, chaos}.
Without adequate representational similarity, the model struggles to generalize across classes, leading to degraded performance on downstream tasks where intra-class variability is not fully captured through augmentations~\cite{chaos_message, understanding_CL, augmentation_overlap}.
Since achieving representational similarity in high-dimensional spaces is challenging~\cite{Parulekar2023InfoNCELP}, a key limitation of augmentation-based self-supervised learning lies in its reliance on handcrafted transformations.
These transformations can distort critical class-level structure, leading to severe performance degradation in complex or heterogeneous data regimes.

Another important limitation of augmentation-based SSL is the inductive bias and feature suppression introduced by both the augmentation process and the optimization objective~\cite{xiao2021what, chen2021intriguing}.
When trained with strong augmentations, models tend to focus on a subset of predictive features, typically those aligned with the enforced invariances, while suppressing other features that may be critical for downstream performance~\cite{robinson2021can}.
This often leads to reliance on a subset of features, harming generalization~\cite{xiao2021what, consistent_representations}.
Moreover, the inductive bias introduced by augmentations acts as a double-edged sword: promoting invariance to certain transformations may benefit some tasks but harm others~\cite{aug_aware, which_features_are_learnt}.
For instance, rotation-based augmentations are commonly used in activity recognition from inertial measurement units to promote robustness across sensor placements~\cite{Tang_cecilia}.
However, they can obscure orientation-dependent features needed to distinguish between fine-grained activities such as standing and sitting, where subtle differences in sensor orientations are important.
These effects are more problematic in signals where the semantic relevance of augmentations varies across tasks.

In this work, we introduce a novel SSL method that generates views by projecting data onto an orthonormal base and an overcomplete frame, and then performs instance discrimination across these spaces.
We demonstrate that the learned representations from instance discrimination lie on distinct manifolds, each shaped by the inherent geometric biases of its corresponding projections.
Building on this observation, we propose to learn mapping functions that transform the original data's space into alternative latent representations.
This mapping enables us to obtain multiple representations using a single encoder augmented with mapping functions to use collections of manifolds.

Our method leverages the inductive bias introduced by projecting data into an orthonormal basis and an overcomplete frame to learn representations.
We summarize our contributions as follows:

\begin{itemize}
\item We propose a novel self-supervised learning method that projects data into an orthonormal basis and an overcomplete frame to perform instance discrimination across these fixed transformations without increasing the data diversity using handcrafted augmentations.

\item We empirically and theoretically show that embeddings from these transformations lie on distinct manifolds shaped by domain-specific geometric biases.
We then jointly leverage these complementary structures to improve performance on downstream tasks.

\item We demonstrate that our method achieves up to 15--20\% performance gains over existing methods on nine datasets across five temporal sequence tasks while using fixed transformations across datasets unlike existing approaches that rely on task-specific augmentations.

\end{itemize}
\section {Method}
\label{sec:Prop_method}

\subsection{Notations}
We use lowercase letters (e.g., \( x \)) to denote scalar quantities, and bold lowercase letters (e.g., \( \boldsymbol{\mathrm{x}} \)) to represent vectors, such as time series, while bold uppercase letters (e.g., \( \boldsymbol{X} \)) are used for matrices.
The parametric function is represented as $f_{\theta}(.)$ where $\theta$ is the parameter.
The discrete Fourier transformation is denoted as $\mathcal{F}(.)$, yielding a complex variable as $\mathcal{F}_{\mathrm{x}}(k) \in \mathbb{C}^k$, where $k$ is the frequency.
The detailed calculations for each operation are given in the Appendix~\ref{appen:proof}.

\subsection{Setup}
We follow the common SSL setup.
Given an unlabeled dataset $\mathcal{D} = \{  (\boldsymbol{\mathrm{x}}_i) \}_{i=1}^K$ where each $\boldsymbol{\mathrm{x}}_i$ is a real-valued sequence of length $L$ with $C$ channels, the goal is to train a learner $f_{\theta}$ that maps inputs to representations $\boldsymbol{h}_i = f_{\theta}(\boldsymbol{\mathrm{x}}_i)$.
To evaluate the learned representations, we train a linear classifier on top of the frozen encoder using a labeled set $\mathcal{D}_l = \{  (\boldsymbol{\mathrm{x}}_i, \boldsymbol{\mathrm{y}}_i) \}_{i=1}^M$ with $M\ll K$ and ${\boldsymbol{\mathrm{y}}_i} \in \{1,\dots, N\} $.

\subsection{Orthonormal Bases and Overcomplete Frames}
Analyzing signals in different domains is useful for detecting desired patterns, as each domain is tailored to capture specific aspects of the data~\cite{cohen_time_frequency, IEEE_old}.
Our method employs the Fourier, \(\mathcal{F}_x(k)\), and the Gabor wavelet transform, \( \mathcal{W}_{\mathrm{x}}(a, b)  \), to construct, respectively, an orthonormal basis (using a tight frame) and an overcomplete frame.
Equation~\ref{eq:cont_transform} defines their corresponding discrete transformations.
\begin{equation}\label{eq:cont_transform}
    \mathcal{F}_{\mathrm{x}}(k) = \frac{1}{\sqrt{L}} \sum_{n}^{} \mathrm{x}(n) \, e^{-j \frac{2\pi}{L} k n}, \hspace{4mm} \mathcal{W}_{\mathrm{x}}(a, b) = \frac{1}{\sqrt{a}} \sum_{n}^{}  \mathrm{x}(n) \, \psi\left( \frac{n - b}{a} \right), 
\end{equation}
where $\psi$ is the Gabor frame.
We use these two transformations as they are complementary.
Specifically, the Fourier transform provides a global overview of the signal's frequency content, while the Gabor wavelet enables localized frequency analysis by zooming on specific time intervals~\cite{wavelet_tour}.


\subsection{Instance Discrimination}
Data representations in the Fourier and Gabor wavelet domains are inherently unique, i.e., distinct samples yield distinct transforms.
Consequently, the instance discrimination task between the views in different domains is well-defined.
Moreover, since these transformations are isometric, the resulting views are not only unique for each sample but also preserve the underlying geometry.
These properties offer significant advantages over existing SSL methods relying on strong augmentations, which may distort samples, causing different classes to appear similar or losing task-relevant information~\cite{InfoMin}.

\begin{figure}[t]
    \centering
    \includegraphics[width=\linewidth]{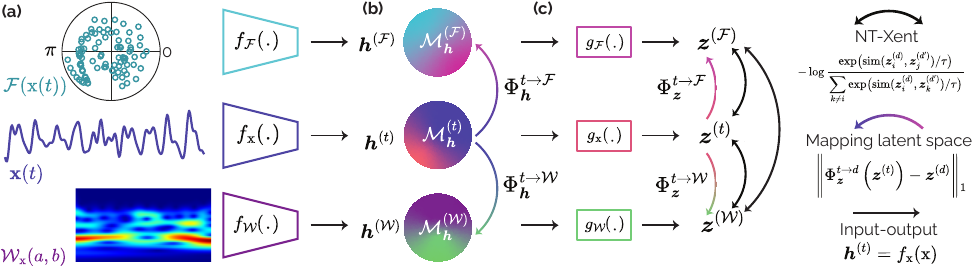}
    \caption{Overview of our method.
    \textbf{(a)} Original data and its transformed versions. The Fourier transformation is given in polar coordinates where we feed to the model magnitude and angle of each harmonic separately.
    \textbf{(b)} Representations \( \boldsymbol{h} \) from each encoder lie on distinct manifolds \( \mathcal{M} \), with latent mappers \( \Phi^{t \rightarrow d}_{\boldsymbol{h}} \) translating across domain-specific spaces. 
    \textbf{(c)} Embeddings \( \boldsymbol{z} \) from each projection, with non-linear mappers used during pre-training to improve predictability across spaces.
    }
    \label{fig:overall}
\end{figure}

We use the normalized temperature-scaled cross-entropy (NT-Xent) loss~\cite{simclr,deep_metric_learning,wu2018unsupervised}, based on cosine similarity, with separate encoders and projection heads for each domain.  
For example, the Fourier branch takes the transformed input \( \mathcal{F}_{\mathrm{x}} \) and produces an embedding via \( \boldsymbol{z}^{\mathcal{F}} = g_{\mathcal{F}}(f_{\mathcal{F}}(\mathcal{F}_{\mathrm{x}})) \), as shown in Figure~\ref{fig:overall}.  
The instance discrimination loss calculated across three domains is defined in Equation~\ref{eq:NT_Xent}.
\begin{equation}\label{eq:NT_Xent}
\ell\bigl(\boldsymbol{z}_i^{(d)}, \boldsymbol{z}_j^{(d')}\bigr)
= -\log 
\frac{
  \exp  \bigl(\mathrm{sim}(\boldsymbol{z}_i^{(d)}, \boldsymbol{z}_j^{(d')})/\tau\bigr)
}{
  \displaystyle\sum_{k \neq i}
    \exp \bigl(\mathrm{sim}(\boldsymbol{z}_i^{(d)}, \boldsymbol{z}_k^{(d')})/\tau\bigr)
}\,, \hspace{2mm} 
\text{where } d \neq d', \hspace{2mm} \text{and} \hspace{2mm} d, d' \in \{t, \mathcal{F}, \mathcal{W}\},
\end{equation}
where $t$ is the time domain.
We compute the final loss as the unweighted sum of all pairwise terms between the time, Fourier, and wavelet domains over a batch of N samples, as shown in Equation~\ref{eq:ID}.
\begin{equation}\label{eq:ID}
\mathcal{L}_{\mathrm{ID}}=\sum_{\substack{d,d'\in\{t,\mathcal{F},\mathcal{W}\}\\d\neq d'}} \frac{1}{2N}\sum_{k=1}^N
\Bigl[
   \ell\bigl(\boldsymbol{z}_{k-1}^{(d)}, \boldsymbol{z}_{k}^{(d')}\bigr)
 + \ell\bigl(\boldsymbol{z}_{k}^{(d)}, \boldsymbol{z}_{k-1}^{(d')}\bigr)
\Bigr]
\end{equation}

\subsubsection{Invariances}
Strong augmentations in instance discrimination can cause encoders to become overly reliant on a subset of features or invariant to some transformations that discard task-relevant information~\cite{xiao2021what, hamidieh2024views}.  
Since the designed augmentations implicitly assume a particular set of representational invariances (e.g., invariance to rotation), and can perform poorly when a downstream task violates this assumption (e.g., distinguishing sitting vs standing)~\cite{sui2024selfsupervised}.
Proposition~\ref{prop:no_invariance_collapse} shows that our method avoids this issue.
\begin{proposition}
\label{prop:no_invariance_collapse}
Let $f_d^{\ast}$ denote an optimal encoder under NT-Xent for domain $d \in \{t,\mathcal{F},\mathcal{W}\}$.  
If for some unintended transformation $W$, the encoder is invariant, i.e., $f_d^{\ast}(W\mathrm{x}) = f_d^{\ast}(\mathrm{x}),$ then for any anchor sample $x$ the NT-Xent loss across domains is lower bounded by the number of negatives.
\[
    \ell\!\left(\boldsymbol{z}_i^{(d)}, \boldsymbol{z}_j^{(d')}\right) \;\;\geq\;\; \log(K+1) > 0,
\]
where $K \geq 1$ is the number of negatives that become near-positives due to the invariance.
\end{proposition}

\begin{proof}
At the NT-Xent optimum, positive pairs align perfectly~\cite{chaos_is_a_ladder, alignment_uniformity},
\[
    f_d^{\ast}(\mathcal{T}(\mathrm{x})) = f_d^{\ast}(\mathrm{x}), \quad \forall d,
\]
for the domain transformations $\mathcal{T} \in \{t,\mathcal{F},\mathcal{W}\}$.  
If $f_d^{\ast}$ is also invariant to $W$, then at least $K \geq 1$ negatives satisfy  
$\boldsymbol{z}_i = f_d^{\ast}(W\mathrm{x}_i)$ with $(f_d^{\ast}(\mathrm{x}), \boldsymbol{z}_i) \approx 1$.  
Even if all other negatives are dissimilar, the denominator of NT-Xent contains at least $K+1$ large terms, yielding
\[
    \ell(\mathrm{x}) \geq -\log \frac{e^{1/\tau}}{(K+1)\, e^{1/\tau}} = \log(K+1).
\]
Thus the loss admits a nontrivial lower bound in the presence of unintended invariance.
\end{proof}

Proposition~\ref{prop:no_invariance_collapse} states that, unlike augmentation-based contrastive learning setups which may encourage spurious invariances~\cite{sui2024selfsupervised, hamidieh2024views}, our method ensures that such invariances cannot minimize the overall optimization objective.  
The detailed derivation of the proposition is provided in Appendix~\ref{appen:proof}.

\subsection{Collections of Manifolds}
Transforming data into well-known bases or overcomplete frames has clear benefits for pattern recognition, but doing this at inference time is costly due to the extra encoders and transformations.
When a transformation operates in the same space as convolutional filters, prior works showed that applying transformations to the filters instead of the input reduces inference cost while still producing diverse representations (e.g., rotating filters instead of rotating the image and reprocessing it~\cite{group_equivariant}).

However, our method transforms data into the complex space where applying equivalent transformations to neural networks is not straightforward.
We therefore propose lightweight latent space mappers \( \Phi \) that transform the original representations into other spaces to leverage their geometry in downstream tasks.
Specifically, we train two mappers, \( \Phi_{\boldsymbol{h}}^{t \rightarrow d}: \mathcal{M}^{(t)}_{\boldsymbol{h}} \rightarrow \mathcal{M}^{(d)}_{\boldsymbol{h}} \), where \( d \in \{\mathcal{F}, \mathcal{W}\} \), to approximate the representations produced by the corresponding domain-specific encoders.

\begin{wrapfigure}[19]{t}{8cm}
\vspace{-3mm}
    \centering
    \includegraphics[width=0.5\columnwidth]{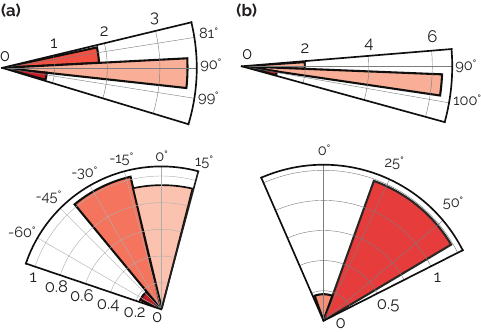}
    \caption{Radial histograms illustrating angle distributions.  
    \textbf{(a)} Top: Angle density of \( \arccos\big( \langle \boldsymbol{h}^{(t)}, \boldsymbol{h}^{(\mathcal{F})} \rangle \big) \);  
    Bottom: Angle density of \( \Delta_{ij} \).  
    \textbf{(b)} Same illustration from the Gabor wavelet \( \boldsymbol{h}^{(\mathcal{W})} \).  
    In both cases, representations of the same samples across domains approach orthogonality, while pairwise angle differences remain widely distributed.
    }
    \label{fig:proof_prop1}
\end{wrapfigure}

Our approach focuses on learning pairwise relationships, such as relative angles between samples across manifolds, rather than mapping individual points between latent spaces.
Therefore, unlike prior methods based on affine latent-space mappings~\cite{latent_space_translation}, we use non-linear mappers.
This is motivated by Proposition~\ref{prop:angle_pairwise}, which shows that in high dimensions, representations of the same sample across spaces can become orthogonal, while pairwise angle variation can span the full range.
This suggests that preserving pairwise geometry is more challenging than aligning individual points.
\begin{proposition}[Angle Concentration vs. Pairwise Spread]\label{prop:angle_pairwise}
Let \( \boldsymbol{h}^{(t)}, \boldsymbol{h}^{(\mathcal{F})} \sim \mathrm{Unif}(S^{d-1}) \), where \( \boldsymbol{h}^{(\mathcal{F})} = f_{\mathcal{F}}(\mathcal{F}(\mathrm{x})) \).  
Although individual samples across latent spaces tend toward orthogonality, the pairwise angular difference \( \Delta_{ij} \) between distinct samples can span the full range up to \( \pi \).
\begin{gather}
\arccos\big( \langle \boldsymbol{h}^{(t)}, \boldsymbol{h}^{(\mathcal{F})} \rangle \big) = \frac{\pi}{2}, \hspace{2mm} \text{while} \hspace{2mm}
\arccos (\langle \boldsymbol{h}_i^{(t)}, \boldsymbol{h}_j^{(t)} \rangle ) - \arccos (\langle \boldsymbol{h}_i^{(\mathcal{F})}, \boldsymbol{h}_j^{(\mathcal{F})} \rangle ) 
= \Delta_{ij} \leq \pi
\end{gather}
\end{proposition}
\begin{proof}
    \begin{equation}
\dim \bigl(\boldsymbol h_i^{(t)\!\perp}\cap\boldsymbol h_j^{(t)\!\perp}\bigr)=d-2\ge1
\;\Longrightarrow\;
\exists\,\boldsymbol h_i^{(\mathcal F)},\boldsymbol h_j^{(\mathcal F)}:
\langle\boldsymbol h_i^{(\mathcal F)},\boldsymbol h_j^{(\mathcal F)}\rangle=\cos\phi,\;\;\forall\phi\in[0,\pi],
    \end{equation}
    Therefore, $|\Delta_{ij}|=\bigl|\theta_{ij}^{(t)}-\theta_{ij}^{(\mathcal F)}\bigr|\le\pi.$
\end{proof}
Figure~\ref{fig:proof_prop1} illustrates Proposition~\ref{prop:angle_pairwise} by showing the angle densities between the same ($i=j$, $\arccos\big( \langle \boldsymbol{h}^{(t)}, \boldsymbol{h}^{(\mathcal{F})} \rangle \big) $) and different samples ($i \neq j$, $|\Delta_{ij}|$) across domains.
We also provide supporting details in Appendix~\ref{appendix:pair_wise_distances}, and the full proof in Appendix~\ref{appen:proof}.
The key point of Proposition~\ref{prop:angle_pairwise} is that preserving near-orthogonality for many pairs does \emph{not} guarantee a global isometry between latent spaces.
Because higher-order relations (e.g., triplet geometry and curvature) can change, the spaces can differ globally despite pairwise near-orthogonality.

In our method, we aim to take advantage of these different geometries, shaped by inductive bias of data and architectures, to improve the performance in the downstream tasks.
Therefore, we employ two mapper functions ($\Phi^{t \rightarrow d}_{\boldsymbol{h}}$) to capture the geometry of latent spaces for other domains.

To improve predictability across latent spaces and reduce estimation error while preserving distinct geometries, we employ lightweight mappers over the embeddings, \( \Phi_{\boldsymbol{z}}^{t \rightarrow d}: \mathcal{M}^{(t)}_{\boldsymbol{z}} \rightarrow \mathcal{M}^{(d)}_{\boldsymbol{z}}, \hspace{2mm} d \in \{\mathcal{F},\mathcal{W} \} \), and optimize them jointly using the loss defined in Equation~\ref{eq:lmap}.

The overall pre-training loss for encoders is the sum of $\mathcal{L}_{\text{map}}$ and $\mathcal{L}_{\text{ID}}$, with no additional weighting.
After pre-training, we freeze the encoders and train the latent space mappers using the same $\mathcal{L}_{\text{map}}$ loss. 
During inference, we only use the main encoder \( f_{\mathrm{x}} \) and the latent space mappers \( \Phi^{t \rightarrow d}_{\boldsymbol{h}} \), excluding all projectors \( g(\cdot) \), auxiliary encoders \( (f_{\mathcal{F}}, f_{\mathcal{W}}) \), and mappers on projected embeddings \( \Phi^{t \rightarrow d}_{\boldsymbol{z}} \).
We provide pseudocode implementation of our method in Appendix~\ref{appen:Algorithm}.

\begin{equation}\label{eq:lmap}
\mathcal{L}_{\text{map}} = 
\frac{1}{N} \left\| \Phi_{\boldsymbol{z}}^{t \rightarrow \mathcal{F}} \left( \boldsymbol{z}^{(t)} \right) - \boldsymbol{z}^{(\mathcal{F})} \right\|_1 
+ 
\frac{1}{N} \left\| \Phi_{\boldsymbol{z}}^{t \rightarrow \mathcal{W}} \left(\boldsymbol{z}^{(t)} \right) - \boldsymbol{z}^{(\mathcal{W})} \right\|_1
\end{equation}

\section{Experiments}
\label{sec:Exp_setup}	

\subsection{Datasets}
\label{ref:datasets}
We conducted experiments on nine datasets across five tasks, including heart rate (HR) estimation from photoplethysmography (PPG), step counting and activity recognition using inertial measurements (IMUs), cardiovascular disease classification from electrocardiogram (ECG) and sleep stage classification from electroencephalography (EEG).
We provide brief descriptions of each dataset below. Additional details, including pre-training and fine-tuning settings are available in Appendix~\ref{appendix:experiments}.
\vspace{-2mm}
\paragraph{\textit{Heart rate}}
We used the IEEE Signal Processing Cup in 2015 (IEEE SPC)~\cite{TROIKA}, and DaLia~\cite{DeepPPG} for PPG-based heart rate prediction from wrist.
We used the leave-one-session-out (LOSO) cross-validation, which evaluates models on subjects/sessions that were not used for training. 
\vspace{-2mm}
\paragraph{\textit{Activity recognition}}
We used HHAR~\cite{hhar}, and USC~\cite{USC} for activity recognition from inertial measurement units from smartphones or wearable devices.
We evaluated the cross-person generalization performance of the models, that is, the model was evaluated on previously unseen subjects. 
\vspace{-2mm}
\paragraph{\textit{Cardiovascular disease (CVD) classification}}
We conducted experiments on China Physiological Signal Challenge (CPSC2018)~\cite{CPSC} and Chapman University, Shaoxing People’s Hospital (Chapman) datasets~\cite{chapman}.
We selected the same four specific leads as in~\cite{Physio_2021} while treating each dataset as a single domain with a small portion of the remaining dataset used for fine-tuning.
We split the dataset for fine-tuning and testing based on patients (each patient's recordings appear in only one set).
\vspace{-2mm}
\paragraph{\textit{Step counting}}
We used the Clemson dataset~\citep{clemson}, which released for pedometer evaluation. 
We conducted experiments using wrist IMUs where labels are available through videos.
\vspace{-2mm}
\paragraph{\textit{Sleep stage classification}}
We used the Sleep-EDF dataset, from PhysioBank~\cite{Physio}, which includes 197 whole-night PSG sleep recordings, where we used a single EEG channel (i.e., Fpz-Cz) with a sampling rate of 100\,Hz, following the same setup as in~\cite{tstcc} while using only 10\% for fine-tuning.

\subsection{Baselines}
\label{ref:baseline}

\paragraph{Fundamentals}
We compare our method to core SSL approaches in the linear evaluation setting~\cite{simclr}.
These include SimCLR~\cite{simclr}, BYOL~\cite{BYOL}, VICReg~\cite{vicreg}, and Barlow Twins~\cite{barlow_twins}.
We also include CLIP~\cite{CLIP}, since data transformations in our method can be interpreted as different data domains.

\paragraph{Temporal sequences}  
We also compare our method with SSL techniques for temporal data, including TS-TCC~\cite{tstcc}, TF-C~\cite{tf_consistent}, simMTM~\cite{dong2023simmtm}, and TS2Vec~\cite{ts2vec}.  
These methods are designed specifically for temporal sequences; for instance, TF-C uses a Fourier encoder during both training and inference, while TS-TCC employs task-specific augmentations with transformer architectures.

\subsection{Implementation}
\label{ref:implementation}
We employed ResNet~\citep{ResNet} with eight blocks~\citep{resnet1d}, as the backbone for the encoder, with the projector consisting of two fully connected layers. 
For latent space mapping, we use a lightweight 1D convolutional that downsamples and reconstructs the input with transposed convolutions, preserving dimensions while enabling non-linear transformations with fewer than 1k parameters.
Similarly, we use two convolutional encoders for the Fourier- and wavelet-transformed inputs.
To ensure a fair comparison, baselines use 384-dimensional encoders as output, while ours uses 128 per encoder.

We train models with a batch size of 1024 for 256 epochs and decay the learning rate using the cosine decay schedule.
After pre-training, we train a single linear layer classifier on features extracted from the frozen pre-trained network.
The models were optimized using Adam~\cite{Adam} with a learning rate of 0.003, while the linear layer was fine-tuned with a learning rate of 0.03.
Reported results are mean and standard deviation values across three independent runs with different seeds. 
For each dataset, we set the Fourier transform length equal to the signal length while excluding negative frequencies.
For the wavelet transform, we use 48 logarithmically spaced scales ranging from 1 to 128 for all datasets.
More details about the implementation and architectures are given in Appendix~\ref{appendix:Implementation_Details}.

\section{Results}
\label{sec:results}
We report the performance of our method against baselines across nine datasets spanning five tasks in Tables~\ref{tab:performance_ppg} to~\ref{tab:performance_ecg_eeg}.
Overall, our approach achieves up to 20–-30\% improvements in some tasks, with an average gain of 10–-15\% across all datasets compared to both general SSL and sequence-specific techniques.
Moreover, unlike prior approaches that employ different augmentations, our method uses the same transformation across tasks without increasing the diversity of the training set.
\begin{table*}[b]
\caption{Performance comparison of our method with other methods for \textit{HR estimation}}
\begin{adjustbox}{width=\columnwidth,center}
\label{tab:performance_ppg}
\renewcommand{\arraystretch}{1.1}
\begin{tabular}{@{}lllllllllll@{}}
\toprule
\multirow{2}{*}{Method} & \multicolumn{3}{l}{IEEE SPC12} & \multicolumn{3}{l}{IEEE SPC22} & \multicolumn{3}{l}{DaLiA} \\ 
\cmidrule(r{15pt}){2-4}  \cmidrule(r{15pt}){5-7}  \cmidrule(r{15pt}){8-10}  
& MAE $\downarrow$ & RMSE $\downarrow$ & $\rho$ $\uparrow$ & MAE $\downarrow$ & RMSE $\downarrow$ & $\rho$  $\uparrow$ & MAE $\downarrow$ & RMSE $\downarrow$ & $\rho$ $\uparrow$ \\
\midrule
\textit{Supervised} & & & &  \\
FCN & 15.13\small$\pm$0.50 & 21.63\small$\pm$0.48 & 52.09\small$\pm$5.43 & 16.57\small$\pm$0.91 & 26.20\small$\pm$0.60 & 55.98\small$\pm$0.78 & 12.45\small$\pm$0.12 & 18.35\small$\pm$0.24 & 56.98\small$\pm$0.78  \\
ResNet & 7.08\small$\pm$0.20 & 13.60\small$\pm$0.38 & 79.60\small$\pm$1.10 & 9.90\small$\pm$1.47 & 16.67\small$\pm$1.60 & 67.58\small$\pm$2.98 & 5.50\small$\pm$0.05 & 10.84\small$\pm$0.03 & 82.10\small$\pm$0.06 \\
\midrule
\textit{Self-Supervised} & & & & \\
SimCLR & 12.42\small$\pm$0.05 & 20.96\small$\pm$0.30 & 73.62\small$\pm$0.52 & 16.41\small$\pm$0.22 & \underline{22.62}\small$\pm$0.39 & 52.16\small$\pm$1.12 & 16.88\small$\pm$0.19 & 22.64\small$\pm$0.22 & 56.37\small$\pm$0.21 & \\
BYOL & 18.71\small$\pm$0.93 & 25.01\small$\pm$1.50 & 69.82\small$\pm$4.36 & 19.44\small$\pm$0.57 & 26.66\small$\pm$0.90 & 46.74\small$\pm$5.02 & 15.59\small$\pm$0.38 & 21.04\small$\pm$0.36 & 57.11\small$\pm$0.06 & \\
VICReg & 13.17\small$\pm$0.82 & 20.38\small$\pm$1.27 & 73.65\small$\pm$0.02 & 16.78\small$\pm$0.47 & 23.10\small$\pm$0.75 & 54.10\small$\pm$1.26 & 15.70\small$\pm$0.15 & 21.83\small$\pm$0.18 & 55.32\small$\pm$0.62
\\
Barlow Twins & 13.22\small$\pm$0.34 & 20.42\small$\pm$0.88 & 67.51\small$\pm$2.01 & 22.08\small$\pm$0.85 & 29.35\small$\pm$0.56 & 35.65\small$\pm$3.40 & 11.87\small$\pm$0.57 & 19.20\small$\pm$0.29 & 62.20\small$\pm$0.40 \\
CLIP & 10.31\small$\pm$0.35 & 17.10\small$\pm$0.30 & 76.00\small$\pm$1.35 & 16.73\small$\pm$1.08 & 26.39\small$\pm$0.44 & 41.59\small$\pm$2.80 & 13.20\small$\pm$0.18 & 20.88\small$\pm$0.22 & 50.42\small$\pm$1.49 \\
TS-TCC & 11.56\small$\pm$0.41 & 18.04\small$\pm$0.66 & \underline{78.38}\small$\pm$1.41 & 16.52\small$\pm$0.34 & 24.86\small$\pm$0.59 & 44.93\small$\pm$3.40 & \underline{10.23}\small$\pm$0.01 & \underline{18.19}\small$\pm$0.04 & \underline{62.77}\small$\pm$0.04 \\

SimMTM & 13.20\small$\pm$0.11 & 18.27\small$\pm$0.22 & 73.78\small$\pm$1.10 & 17.03\small$\pm$0.63 & 25.18\small$\pm$0.98 & 51.33\small$\pm$2.62 & 13.61\small$\pm$0.05 & 20.12\small$\pm$0.07 & 55.47\small$\pm$0.09 \\

TF-C & 12.10\small$\pm$0.15 & 20.12\small$\pm$0.37 & 66.01\small$\pm$1.14 & \underline{14.12}\small$\pm$0.36 & 22.86\small$\pm$0.44 & \underline{52.74}\small$\pm$1.40 & 16.15\small$\pm$0.43 & 23.47\small$\pm$0.13 & 50.12\small$\pm$1.45 \\
TS2Vec & \underline{9.80}\small$\pm$0.49 & \underline{16.64}\small$\pm$0.55 & 75.30\small$\pm$1.10 & 24.57\small$\pm$0.30 & 24.83\small$\pm$0.16 & 50.40\small$\pm$2.31 & 12.65\small$\pm$0.32 & 20.04\small$\pm$0.33 & 58.17\small$\pm$0.53 \\
Ours & \textbf{8.84}\small$\pm$0.50 & \textbf{14.37}\small$\pm$0.95 & \textbf{82.67}\small$\pm$1.30 & \textbf{14.06}\small$\pm$1.09 & \textbf{21.48}\small$\pm$2.01 & \textbf{54.88}\small$\pm$1.89 & \textbf{9.13}\small$\pm$0.20 & \textbf{16.92}\small$\pm$0.56 & \textbf{63.72}\small$\pm$0.06 & \\
\bottomrule
\end{tabular}
\end{adjustbox}
\end{table*}

\begin{table*}[t]
\caption{Performance comparison of our method with other techniques for \textit{Activity} and \textit{Step}}
\begin{adjustbox}{width=\columnwidth,center}
\label{tab:performance_imu}
\renewcommand{\arraystretch}{1.1}
\begin{tabular}{@{}lllllllllll@{}}
\toprule
\multirow{2}{*}{Method} & \multicolumn{3}{l}{HHAR} & \multicolumn{3}{l}{USC} & \multicolumn{3}{l}{Clemson} \\ 
\cmidrule(r{15pt}){2-4}  \cmidrule(r{15pt}){5-7}  \cmidrule(r{15pt}){8-10}  
& Acc $\uparrow$ & W-F1 $\uparrow$ & F1 $\uparrow$ & Acc $\uparrow$ & W-F1 $\uparrow$ & F1 $\uparrow$ & MAPE $\downarrow$ & MAE $\downarrow$ & RMSE $\downarrow$  \\
\midrule
\textit{Supervised} & & & &  \\
FCN & 74.21\small$\pm$1.56 & 72.88\small$\pm$2.06 & 71.58\small$\pm$1.81 & 48.87\small$\pm$0.74 & 46.02\small$\pm$0.95 & 45.33\small$\pm$0.82 & 5.02\small$\pm$0.26 & 2.86\small$\pm$0.15 & 4.05\small$\pm$0.13   \\
ResNet & 69.85\small$\pm$2.32 & 68.61\small$\pm$2.81 & 67.29\small$\pm$2.52 & 52.17\small$\pm$1.22 & 49.38\small$\pm$0.84 & 48.01\small$\pm$1.22 & 6.55\small$\pm$2.37 & 3.78\small$\pm$1.44 & 5.04\small$\pm$1.43 \\
\midrule
\textit{Self-Supervised} & & & & \\
SimCLR & 40.55\small$\pm$0.62 & 39.21\small$\pm$0.64 & 39.41\small$\pm$0.66 & 29.16\small$\pm$0.69 & 29.02\small$\pm$0.67 & 28.99\small$\pm$0.79 & 8.70\small$\pm$0.22 & 4.36\small$\pm$0.13 & 6.30\small$\pm$0.24 \\
BYOL & 49.64\small$\pm$2.48 & 48.63\small$\pm$2.75 & 48.02\small$\pm$2.59 & 28.40\small$\pm$1.23 & 28.23\small$\pm$1.42 & 28.23\small$\pm$0.96 & 9.35\small$\pm$0.19 & 4.72\small$\pm$0.12 & 6.79\small$\pm$0.24 & \\
VICReg & 38.05\small$\pm$3.01 & 37.12\small$\pm$2.66 & 37.38\small$\pm$3.02 & 23.75\small$\pm$1.00 & 23.16\small$\pm$1.03 & 22.92\small$\pm$1.21 & 10.87\small$\pm$0.61 & 5.47\small$\pm$0.35 & 7.78\small$\pm$0.14
\\
Barlow Twins & 38.97\small$\pm$0.65 & 37.75\small$\pm$1.00 & 38.21\small$\pm$1.12 & 27.24\small$\pm$0.19 & 26.84\small$\pm$0.20 & 26.25\small$\pm$0.77 & 9.89\small$\pm$0.35 & 4.95\small$\pm$0.15 & 7.03\small$\pm$0.21 \\
CLIP & 43.78\small$\pm$0.89 & 42.53\small$\pm$0.90 & 43.07\small$\pm$0.98 & 25.55\small$\pm$0.63 & 25.78\small$\pm$1.25 & 25.17\small$\pm$0.75 & 8.52\small$\pm$0.46 & 4.26\small$\pm$0.23 & 6.73\small$\pm$0.63 \\
TS-TCC & \underline{68.56}\small$\pm$1.19 & \underline{66.90}\small$\pm$1.22 & \underline{68.10}\small$\pm$1.30 & 33.61\small$\pm$0.72 & \underline{33.11}\small$\pm$1.09 & 33.91\small$\pm$0.79 & \underline{5.61}\small$\pm$0.15 & \underline{2.70}\small$\pm$0.06 & \underline{4.69}\small$\pm$0.38 \\

SimMTM & 44.78\small$\pm$0.62 & 42.48\small$\pm$0.37 & 43.60\small$\pm$0.62 & 22.34\small$\pm$0.28 & 25.68\small$\pm$0.41 & 29.72\small$\pm$1.78 & 8.77\small$\pm$0.18 & 4.61\small$\pm$0.32 & 6.90\small$\pm$0.18 \\

TF-C & 31.13\small$\pm$0.42 & 30.57\small$\pm$0.40 & 31.00\small$\pm$0.31 & 30.78\small$\pm$0.39 & 28.16\small$\pm$0.23 & 30.82\small$\pm$1.41 & 12.47\small$\pm$0.72 & 6.31\small$\pm$0.37 & 7.93\small$\pm$0.30 \\
TS2Vec & 67.13\small$\pm$0.11 & 65.56\small$\pm$0.21 & 64.13\small$\pm$0.21 & \underline{35.40}\small$\pm$0.96 & 32.17\small$\pm$1.26 & \underline{35.47}\small$\pm$1.42 & 5.92\small$\pm$0.93 & 3.01\small$\pm$0.28 & 5.02\small$\pm$0.42 \\
Ours & \textbf{70.67}\small$\pm$0.06 & \textbf{67.74}\small$\pm$0.29 & \textbf{68.79}\small$\pm$0.25 & \textbf{52.21}\small$\pm$1.09 & \textbf{48.64}\small$\pm$1.52 & \textbf{48.22}\small$\pm$1.11 & \textbf{5.16}\small$\pm$0.44 & \textbf{2.50}\small$\pm$0.13 & \textbf{4.65}\small$\pm$0.45 & \\
\bottomrule
\end{tabular}
\end{adjustbox}
\end{table*}

\begin{table*}[t]
\caption{Performance comparison of our method with other techniques for \textit{CVD} and \textit{Sleep}}
\begin{adjustbox}{width=\columnwidth,center}
\label{tab:performance_ecg_eeg}
\renewcommand{\arraystretch}{1.1}
\begin{tabular}{@{}lllllllllll@{}}
\toprule
\multirow{2}{*}{Method} & \multicolumn{3}{l}{Chapman} & \multicolumn{3}{l}{CPSC} & \multicolumn{3}{l}{Sleep} \\ 
\cmidrule(r{15pt}){2-4}  \cmidrule(r{15pt}){5-7}  \cmidrule(r{15pt}){8-10}  
& Acc $\uparrow$ & AUC $\uparrow$ & F1 $\uparrow$ & Acc $\uparrow$ & AUC $\uparrow$ & F1 $\uparrow$ & Acc $\uparrow$ & W-F1 $\uparrow$ & Kappa $\uparrow$  \\
\midrule
\textit{Supervised} & & & &  \\
FCN & 84.63\small$\pm$2.13 & 95.40\small$\pm$0.57 & 82.41\small$\pm$2.40 & 63.64\small$\pm$1.12 & 91.30\small$\pm$0.02 & 60.43\small$\pm$1.04 & 71.98\small$\pm$0.86 & 63.33\small$\pm$0.84 & 62.01\small$\pm$1.30  \\
ResNet & 93.16\small$\pm$0.41 & 98.59\small$\pm$0.05 & 92.02\small$\pm$0.42 & 75.21\small$\pm$1.73 & 95.02\small$\pm$0.03 & 71.70\small$\pm$1.90 & 76.94\small$\pm$0.97 & 67.52\small$\pm$1.95 & 69.14\small$\pm$0.61 \\
\midrule
\textit{Self-Supervised} & & & & \\
SimCLR & 75.28\small$\pm$0.57 & 93.55\small$\pm$0.25 & 74.04\small$\pm$0.50 & 50.10\small$\pm$0.41 & \textbf{87.20}\small$\pm$0.07 & 50.10\small$\pm$0.24 & 72.45\small$\pm$2.32 & 58.93\small$\pm$1.59 & 59.47\small$\pm$3.20 & \\
BYOL & 76.08\small$\pm$0.40 & 93.54\small$\pm$0.18 & 74.80\small$\pm$0.45 & \underline{51.90}\small$\pm$0.30 & 87.05\small$\pm$0.22 & \underline{50.89}\small$\pm$0.38 & 70.77\small$\pm$0.27 & 58.23\small$\pm$0.55 & 55.90\small$\pm$1.20 \\
VICReg & 70.10\small$\pm$1.90 & 89.35\small$\pm$0.93 & 67.84\small$\pm$1.79 & 46.21\small$\pm$1.29 & 84.70\small$\pm$0.50 & 42.51\small$\pm$0.96 & 68.72\small$\pm$1.03 & 57.24\small$\pm$1.04 & 57.13\small$\pm$1.42
\\
Barlow Twins & 72.43\small$\pm$1.45 & 91.17\small$\pm$0.60 & 70.42\small$\pm$1.53 & 48.67\small$\pm$0.51 & 85.78\small$\pm$0.19 & 44.57\small$\pm$0.53 & 70.10\small$\pm$0.62 & 57.72\small$\pm$0.81 & 57.88\small$\pm$0.82 \\
CLIP & 82.98\small$\pm$0.96 & 95.15\small$\pm$0.42 & 81.00\small$\pm$1.03 & 50.01\small$\pm$0.89 & 86.40\small$\pm$0.32 & 47.99\small$\pm$0.89 & 73.16\small$\pm$0.81 & 62.06\small$\pm$0.91 & 63.75\small$\pm$1.23 \\
TS-TCC & 73.50\small$\pm$0.55 & 90.65\small$\pm$0.07 & 71.10\small$\pm$0.57 & 51.59\small$\pm$1.22 & 86.32\small$\pm$0.16 & 50.27\small$\pm$1.32 & 62.80\small$\pm$1.13 & 52.43\small$\pm$1.05 & 48.98\small$\pm$1.68 \\

SimMTM & 84.29\small$\pm$1.29 & 95.87\small$\pm$0.18 & 83.31\small$\pm$1.25 & 51.70\small$\pm$0.23 & 87.08\small$\pm$0.21 & 50.62\small$\pm$0.55 & \underline{74.69}\small$\pm$1.84 & \underline{63.53}\small$\pm$1.21 & \underline{65.31}\small$\pm$2.76 \\

TF-C & \underline{85.84}\small$\pm$0.39 & \underline{96.10}\small$\pm$0.10 & 84.71\small$\pm$0.40 & 47.86\small$\pm$0.69 & 86.27\small$\pm$0.05 & 45.42\small$\pm$0.66 & 64.50\small$\pm$1.80 & 56.77\small$\pm$2.21 & 52.61\small$\pm$2.41 \\
TS2Vec & 78.87\small$\pm$1.03 & 90.23\small$\pm$0.24 & 81.32\small$\pm$0.47 & 48.73\small$\pm$0.85 & 85.49\small$\pm$0.37 & 46.57\small$\pm$1.10 & 65.71\small$\pm$1.06 & 55.32\small$\pm$1.77 & 56.81\small$\pm$1.90 \\
Ours & \textbf{87.21}\small$\pm$0.80 & \textbf{96.50}\small$\pm$0.21 & \textbf{85.30}\small$\pm$0.98 & \textbf{52.10}\small$\pm$0.90 & \underline{87.11}\small$\pm$0.40 & \textbf{51.26}\small$\pm$1.18 & \textbf{77.30}\small$\pm$1.04 & \textbf{68.05}\small$\pm$0.86 & \textbf{69.16}\small$\pm$1.32 & \\
\bottomrule
\end{tabular}
\end{adjustbox}
\end{table*}

Our main results show that our method outperforms previous techniques by a significant margin in several datasets.
To further investigate, we explore whether prior methods can close this gap with comparable modifications.
Specifically, we ask the following questions.
\begin{enumerate}[leftmargin=20pt, itemsep=0.5pt, topsep=0pt]
\item Our model includes lightweight mappers with additional backbones. 
We ask if prior methods can match our performance by increasing backbone capacity, that is, by brute-force scaling.

\item Our model employs instance discrimination task with more than one view. 
Thus, we also ask if existing techniques with multiple positive views can match our performance.
\end{enumerate}

\begin{wrapfigure}[19]{t}{7.5cm}
\vspace{-3mm}
    \centering
    \includegraphics[width=0.5\columnwidth]{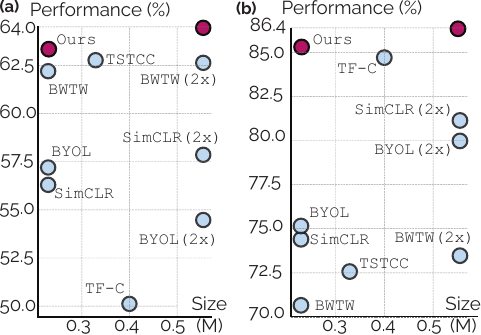}
    \caption{Performance of our method on DaLiA \textbf{(a)} using correlation ($\rho$) and on Chapman \textbf{(b)} using F1 score, compared to other self-supervised learning techniques. Barlow Twins is abbreviated as {\footnotesize (\texttt{BWTW})}, and backbone sizes are shown in millions of parameters.
    The red circle in each plot denotes our method, which achieves higher performance with fewer parameters.
    }
    \label{fig:scale}
\end{wrapfigure}

To answer the first question, we increased the backbone capacity of previous fundamental techniques, especially SimCLR, BYOL and Barlow Twins as they represent unique approaches, by 2x compared to our method by adding additional residual blocks to the backbone.
Results are given in Figure~\ref{fig:scale}.
We omit the error bars for figures where the standard deviations were negligible relative to the means.
As shown in the results, simply increasing model size does not close the performance gap.
In fact, larger models often exhibit stable or decreased performance, particularly in heterogeneous low-data regimes, likely due to overfitting even with strong augmentations.
In contrast, our method consistently outperforms them across all tasks, achieving 10--15\% higher performance with approximately half the number of parameters.

One observation from this comparison is that TF-C performs worse than others for some datasets, despite using two encoders, one for the time and one for the Fourier domain.
We hypothesize that this performance drop may stem from the strong augmentations applied in TF-C framework, which changes the magnitude of frequency components and degrade representations for noisy signals.

\begin{figure}[ht]
  \centering
  \begin{subfigure}[t]{0.32\textwidth}
    \centering
    \includegraphics[width=\linewidth]{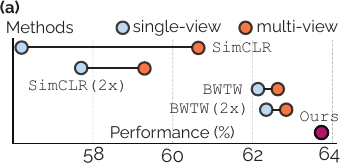}
    \label{fig:view_single}
  \end{subfigure}\hfill
  \begin{subfigure}[t]{0.32\textwidth}
    \centering
    \includegraphics[width=\linewidth]{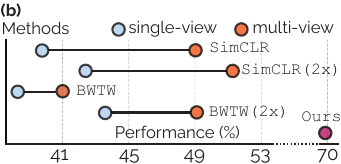}
    \label{fig:view_multi}
  \end{subfigure}\hfill
  \begin{subfigure}[t]{0.32\textwidth}
    \centering
    \includegraphics[width=\linewidth]{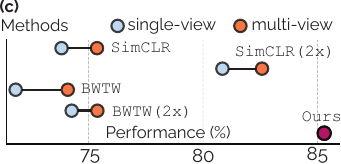}
    \label{fig:view_third}
  \end{subfigure}
  \caption{Performance comparison across datasets, \textbf{(a)} DaLiA ($\rho$), \textbf{(b)} HHAR (Acc), and \textbf{(c)} Chapman (F1), when adding a third view for instance discrimination similar to our method.
  While prior methods benefit from the additional view, their performance still lags behind our approach by a large margin.
  }
  \label{fig:multi_view_three}
\end{figure}

For the second question, we generated a third view using a task-specific random augmentation and applied the instance discrimination loss across all views, following the same setup as our method.
Figure~\ref{fig:multi_view_three} presents the results, including comparisons with larger backbone for prior techniques.

These results show that adding a third view improves performance for prior methods, but they still fall short of ours by about 10\%.  
Notably, introducing a third view increases data diversity for these methods, while in our case, using a third transformation does not.  
This empirical evidence highlights the data-hungry nature of prior SSL approaches and the efficiency of our method further. 

\subsection{Ablation Experiments}
Our method consists of multiple components, so we conduct many ablation studies to assess the contribution of each.
First, we evaluate the impact of each data transformation by selectively removing them individually.
We exclude the orthonormal basis transformation (w/o OB) and retain only the overcomplete frame, then exclude the overcomplete frame transformation (w/o OF) and retain only the orthonormal basis in the method.
We present the results across tasks in Tables~\ref{tab:performance_ppg_ablation},~\ref{tab:performance_activity_ablation}, and~\ref{tab:performance_ecg_eeg_ablation}.
\begin{table*}[h]
\centering
\caption{Ablation on proposed method in \textit{PPG} datasets for HR estimation}
\begin{adjustbox}{width=\columnwidth,center}
\label{tab:performance_ppg_ablation}
\renewcommand{\arraystretch}{1.1}
\begin{tabular}{@{}lllllllllll@{}}
\toprule
\multirow{2}{*}{Method} & \multicolumn{3}{l}{IEEE SPC12} & \multicolumn{3}{l}{IEEE SPC22} & \multicolumn{3}{l}{DaLiA$_{PPG}$} \\ 
\cmidrule(r{15pt}){2-4}  \cmidrule(r{15pt}){5-7}  \cmidrule(r{15pt}){8-10} 
& MAE $\downarrow$ & RMSE $\downarrow$ & $\rho$ $\uparrow$ & MAE $\downarrow$ & RMSE $\downarrow$ & $\rho$ $\uparrow$ & MAE $\downarrow$ & RMSE $\downarrow$ & $\rho$ $\uparrow$ \\
\midrule
Ours & \textbf{8.84}\small$\pm$0.50 & \textbf{14.37}\small$\pm$0.95 & \textbf{82.67}\small$\pm$1.30 & \textbf{14.06}\small$\pm$1.09 & \textbf{21.48}\small$\pm$2.01 & \textbf{54.88}\small$\pm$1.89 & 9.13\small$\pm$0.20 & 16.92\small$\pm$0.56 & 63.72\small$\pm$0.06 & \\
w/o OB & 11.23 \small(\textcolor{WildStrawberry}{+2.39})  & 17.26 \small(\textcolor{WildStrawberry}{+2.89}) & 75.99 \small(\textcolor{WildStrawberry}{-6.68}) & 15.88 \small(\textcolor{WildStrawberry}{+1.82}) & 24.82 \small(\textcolor{WildStrawberry}{+3.34}) & 53.96 \small(\textcolor{WildStrawberry}{-0.92}) & \textbf{8.87} \small(\textcolor{Green}{-0.26}) & \textbf{16.13} \small(\textcolor{Green}{-0.79}) & \textbf{65.34} \small(\textcolor{Green}{+1.62})  \\
w/o OF & 9.80 \small (\textcolor{WildStrawberry}{+0.96}) & 15.18 \small(\textcolor{WildStrawberry}{+0.81}) & 81.19 \small(\textcolor{WildStrawberry}{-1.48}) & 15.33 \small(\textcolor{WildStrawberry}{+1.27}) & 24.29 \small(\textcolor{WildStrawberry}{+2.81}) & 53.05 \small(\textcolor{WildStrawberry}{-1.83}) & 9.60 \small(\textcolor{WildStrawberry}{+0.47}) & 17.31 \small(\textcolor{WildStrawberry}{+0.39}) & 63.06 \small(\textcolor{WildStrawberry}{-0.66}) \\
\midrule
w/o $\Phi^{t \rightarrow d}_{\boldsymbol{h}} $ & 9.62 \small (\textcolor{WildStrawberry}{+0.78}) & 15.46 \small(\textcolor{WildStrawberry}{+1.09}) & 80.21 \small(\textcolor{WildStrawberry}{-2.46}) & 14.16 \small(\textcolor{WildStrawberry}{+0.10}) & 22.46 \small(\textcolor{WildStrawberry}{+0.98}) & 56.23 \small(\textcolor{WildStrawberry}{+2.35}) & 8.90 \small(\textcolor{Green}{-0.23}) & 16.27 \small(\textcolor{Green}{-0.65}) & 64.65 \small(\textcolor{Green}{+0.93}) \\
\bottomrule
\end{tabular}
\end{adjustbox}
\end{table*}

\begin{table*}[t]
\centering
\caption{Ablation on proposed method in \textit{IMU} datasets for Activity and Step}
\begin{adjustbox}{width=\columnwidth,center}
\label{tab:performance_activity_ablation}
\renewcommand{\arraystretch}{1.1}
\begin{tabular}{@{}lllllllllll@{}}
\toprule
\multirow{2}{*}{Method} & \multicolumn{3}{l}{HHAR} & \multicolumn{3}{l}{USC} & \multicolumn{3}{l}{Clemson} \\ 
\cmidrule(r{15pt}){2-4}  \cmidrule(r{15pt}){5-7}  \cmidrule(r{15pt}){8-10} 
& Acc $\uparrow$ & W-F1 $\uparrow$ & F1 $\uparrow$ & Acc $\uparrow$ & W-F1 $\uparrow$ & F1 $\uparrow$ & MAPE $\downarrow$ & MAE $\downarrow$ & RMSE $\downarrow$ \\
\midrule
Ours & \textbf{70.67}\small$\pm$0.06 & \textbf{67.74}\small$\pm$0.29 & \textbf{68.79}\small$\pm$0.25 & \textbf{52.21}\small$\pm$1.09 & \textbf{48.64}\small$\pm$1.52 & \textbf{48.22}\small$\pm$1.11 & \textbf{5.16}\small$\pm$0.44 & \textbf{2.50}\small$\pm$0.23 & \textbf{4.65}\small$\pm$0.50 & \\
w/o OB & 67.45 \small (\textcolor{WildStrawberry}{-3.22})  & 65.10 \small (\textcolor{WildStrawberry}{-2.64}) & 86.50 \small (\textcolor{WildStrawberry}{-2.09}) & 44.70 \small(\textcolor{WildStrawberry}{-7.51}) & 41.20 \small(\textcolor{WildStrawberry}{-7.44}) & 41.95 \small(\textcolor{WildStrawberry}{-6.27}) & 7.50 \small(\textcolor{WildStrawberry}{+2.34}) & 3.65 \small(\textcolor{WildStrawberry}{+1.15}) & 5.79 \small(\textcolor{WildStrawberry}{+1.14})  \\
w/o OF & 69.74 \small(\textcolor{WildStrawberry}{-0.93}) & 67.62 \small(\textcolor{WildStrawberry}{-1.12})  & 68.01 \small(\textcolor{WildStrawberry}{-0.78}) & 49.06 \small(\textcolor{WildStrawberry}{-3.15}) & 45.17 \small(\textcolor{WildStrawberry}{-3.47}) & 45.61 \small(\textcolor{WildStrawberry}{-2.61}) & 5.59 \small(\textcolor{WildStrawberry}{+0.43}) & 2.72 \small(\textcolor{WildStrawberry}{+0.22}) & 5.07 \small(\textcolor{WildStrawberry}{+0.42}) \\
\midrule
w/o $\Phi^{t \rightarrow d}_{\boldsymbol{h}} $ & 67.11 \small(\textcolor{WildStrawberry}{-3.56}) & 66.24 \small(\textcolor{WildStrawberry}{-1.50}) & 67.05 \small(\textcolor{WildStrawberry}{-1.74}) & 51.26 \small(\textcolor{WildStrawberry}{-0.95}) & 47.81 \small(\textcolor{WildStrawberry}{-0.83}) & 47.70 \small(\textcolor{WildStrawberry}{-0.52}) & 5.20 \small(\textcolor{WildStrawberry}{+0.04}) & 2.45 \small(\textcolor{Green}{-0.05}) & 4.53 \small(\textcolor{Green}{-0.12}) \\
\bottomrule
\end{tabular}
\end{adjustbox}
\end{table*}

\begin{table*}[t]
\centering
\caption{Ablation on proposed method in \textit{ECG} and \textit{EEG} datasets for CVD and Sleep}
\begin{adjustbox}{width=\columnwidth,center}
\label{tab:performance_ecg_eeg_ablation}
\renewcommand{\arraystretch}{0.9}
\begin{tabular}{@{}lllllllllll@{}}
\toprule
\multirow{2}{*}{Method} & \multicolumn{3}{l}{Chapman} & \multicolumn{3}{l}{CPSC} & \multicolumn{3}{l}{Sleep} \\ 
\cmidrule(r{15pt}){2-4}  \cmidrule(r{15pt}){5-7}  \cmidrule(r{15pt}){8-10} 
& Acc $\uparrow$ & F1 $\uparrow$ & AUC $\uparrow$ & Acc $\uparrow$ & F1 $\uparrow$ & AUC $\uparrow$ & Acc $\uparrow$ & F1 $\uparrow$ & Kappa ($\kappa$) $\uparrow$ \\
\midrule
Ours & \textbf{87.21}\small$\pm$0.80 & 96.50\small$\pm$0.21 & 85.30\small$\pm$0.98 & 52.10\small$\pm$0.90 & 87.11\small$\pm$0.40 & 51.26\small$\pm$1.18 & 77.30\small$\pm$1.04 & 68.05\small$\pm$0.86 & 69.16\small$\pm$1.32 \\ 
w/o OB & 86.79 \small(\textcolor{WildStrawberry}{-0.42})  & \textbf{96.51} \small(\textcolor{Green}{+0.01}) & \textbf{85.90} \small(\textcolor{Green}{+0.60}) & \textbf{58.81} \small(\textcolor{Green}{+6.71}) & \textbf{89.35} \small(\textcolor{Green}{+2.24}) & \textbf{55.91} \small(\textcolor{Green}{+
4.65}) & \textbf{81.15} \small(\textcolor{Green}{+3.85}) & \textbf{70.73} \small(\textcolor{Green}{+2.68}) & \textbf{74.37} \small(\textcolor{Green}{+5.21})  \\
w/o OF & 81.82 \small(\textcolor{WildStrawberry}{-5.39}) & 94.15 \small(\textcolor{WildStrawberry}{-2.35})  & 79.26 \small(\textcolor{WildStrawberry}{-6.04}) & 45.83 \small(\textcolor{WildStrawberry}{-6.27}) & 84.47 \small(\textcolor{WildStrawberry}{-2.64}) & 44.08 \small(\textcolor{WildStrawberry}{-7.18}) & 73.91 \small(\textcolor{WildStrawberry}{-3.39}) & 63.98 \small(\textcolor{WildStrawberry}{-4.07}) & 64.75 \small(\textcolor{WildStrawberry}{-4.41}) \\
\midrule
w/o $\Phi^{t \rightarrow d}_{\boldsymbol{h}} $ & 84.98 \small(\textcolor{WildStrawberry}{-2.23}) & 95.70 \small(\textcolor{WildStrawberry}{-0.80}) & 82.69 \small(\textcolor{WildStrawberry}{-2.61}) &  51.44 \small(\textcolor{WildStrawberry}{-0.66}) & 86.59 \small(\textcolor{WildStrawberry}{-0.52}) & 49.63 \small(\textcolor{WildStrawberry}{-1.63}) & 79.30 \small(\textcolor{Green}{+2.00}) & 68.40 \small(\textcolor{Green}{+0.35}) & 71.67 \small(\textcolor{WildStrawberry}{-2.51}) \\
\bottomrule
\end{tabular}
\end{adjustbox}
\end{table*}

These results show that the best performance is mostly achieved when both transformations are used together, though the degree of performance change varies across tasks.
One interesting result from our ablation study is that using only the Gabor wavelet transform sometimes outperforms the combination with Fourier transformation.
Specifically, we observe that the Gabor wavelet is more effective for tasks involving sudden signal changes (e.g., abnormal heart rhythms, neural activity during sleep), while the Fourier transform better captures global structures such as periodic patterns in heart rate or step counting from inertial measurements.
This empirical evidence supports our motivation for using complementary, principled transformations to jointly capture diverse signal characteristics.

Second, we retain both transformations and their NT-Xent loss but exclude the latent space mappers \( \Phi^{t \rightarrow d }_{\boldsymbol{h}} \) during inference, performing linear probing solely on the original latent space \( \boldsymbol{h}^{(t)} \).
Results are reported in the same tables with the first ablation experiment under "w/o \( \Phi^{t \rightarrow d}_{\boldsymbol{h}} \)".
As the results indicate, removing the mappers from our method degrades performance compared to the best case.

We conducted additional ablations to evaluate the performance of latent space mappers $\Phi_{\boldsymbol{h}}^{t \rightarrow d}$ and to assess the impact of embedding mappers $\Phi_{\boldsymbol{z}}^{t \rightarrow d}$ on the performance.
Results are given in Appendix~\ref{appendix:additional_experiments}.

\subsection{Discussion of results}

\paragraph{Do we need data augmentations?}
Data augmentations are widely viewed as essential for learning representations from unlabeled data~\cite{simclr, generative_or_contrastive}.
A celebrated theory, InfoMin~\cite{InfoMin}, argues that effective augmentations reduce mutual information between views while retaining task-relevant features.
However, our method does not rely on asymmetric views but instead uses unitary transformations that preserve all information.
Our results suggest that asymmetric views are not essential for SSL.

A recent study~\cite{chaos} highlights the importance of strong augmentations in instance discrimination by showing that augmentations can cause different intra-class samples to align when their augmented views overlap (i.e., two different cars appear similar when both are cropped to show only the wheels).
However, this explanation does not extend to our method.
Since our transformations are unitary, the views preserve the structure of the original sample without introducing overlaps across instances.
Our findings show that instance-discrimination–based self-supervised learning can succeed without relying on hand-crafted strong augmentations for temporal signals.
This opens the door for future work to test whether the same holds in other modalities, such as images and audio.

\paragraph{Implicit bias}
An interesting finding from our experiments is that, although all encoders are trained with the same loss at the same time and without stop-gradient operations (commonly used to prevent collapse), the geometries of the learned representation differ significantly across encoders.
We attribute this to the applied data transformations, which introduce strong implicit biases that shape each latent space in ways that emphasize different characteristics, such as global or local features.

\section{Limitations}
\label{sec:limitations}
While our work demonstrates that instance discrimination-based SSL can be effective without aggressive data augmentations, we do not provide a theoretical explanation for the observed performance gains.  
We hypothesize that the representational improvements arise from implicit biases of the transformations, though this remains unverified by formal theoretical analysis.  

In terms of scope, our experiments primarily focused on classification tasks, as this setting involves a wide variety of augmentations, including task-specific ones~\cite{demirel2023chaos}, making it a natural testbed for our augmentation-free approach.  
Nevertheless, since our method leverages both global (FFT) and local (Gabor) representations for representation learning, it also holds potential for forecasting tasks where capturing both long-range and localized temporal patterns is critical.  
Exploring this direction in both self-supervised and supervised paradigms remains an important avenue for future work. 

Finally, while our method applies Fourier and wavelet transformations, it is worth noting that although the Fourier transform is computationally efficient~\cite{fft}, computing wavelet coefficients is expensive.  
In our experiments, we mitigated this by caching the coefficients after a one-time computation.  
We quantitatively evaluate the computational overhead of our method and other techniques in Appendix~\ref{sec:computational_overhead}, showing that our approach maintains competitive runtime compared to state-of-the-art works.  


\clearpage
\section{Related Work}
\label{sec:related_work}
\paragraph{Data augmentation in SSL}
Learning representations through data transformations dates back to early self-supervised learning methods, which introduced pretext tasks that reformulated the problem into a supervised one, such as predicting image rotations~\cite{rotation_prediction} or spatial contexts~\cite{Context_prediction,Context_encoders}.
More recently, data augmentations have become a central component of SSL, with stronger transformations applied to boost sample diversity~\cite{simclr, chaos, tstcc}.
However, learning representations with strong augmentations introduce new challenges~\cite{which_features_are_learnt, Rethinking_augmentation}.
Mainly, strong augmentations can alter the label of a sample~\cite{stronger_stronger}, leading to model collapse~\cite{InfoMin}, and may suppress informative features~\cite{which_features_are_learnt}.
This often causes models to rely on a subset of features aligned with augmentation-induced invariances, potentially ignoring others that are critical for downstream performance~\cite{robinson2021can, consistent_representations, which_features_are_learnt}.
Motivated by these, we propose a method that replaces data augmentations with principled, well-understood transformations.
Instead of relying on task-specific, hand-tuned augmentations, our approach performs instance discrimination using views generated from principled transformations.

\paragraph{Mapping between latent spaces}
Mapping between different representations is a growing area in deep learning as it enables the use of pre or co-trained models across domains~\cite{barak_boaz_stitch, similarity_and_matching}.
For instance, authors in~\cite{moschella2023relative} proposed a zero-shot communication method between latent spaces by projecting them into a shared relative space, constructed from pairwise distances between anchor points.
However, this approach relies on anchor points to capture the structure of each latent space.
More recent work focuses on directly translating between latent spaces using affine or orthogonal transformations~\cite{maiorca2023latent}.
These methods are inspired by Procrustes analysis for latent space alignment~\cite{mahadevan_ICML, mahadevan_2}, which has been applied in language processing~\cite{enriching, smith2017offline}.
Yet, orthogonal transformations preserve inner products and therefore cannot capture the full representational differences learned by distinct encoders---an aspect central to our approach.
Instead, our method estimates the geometry of the target latent space, shaped by domain-specific transformation biases, and enables its use in downstream tasks.

\paragraph{Multiview contrastive learning}
Learning representations with instance discrimination using multiple views without strong augmentations was proposed early in SSL~\cite{contrastive_multiview,multiview_two, dual_contrastive_learning}.
Earlier methods~\cite{contrastive_multiview} used different channels (e.g., \(L\) and \(ab\) from an RGB image) as views, which act as implicit augmentations and may bias the model toward certain features.  
In contrast, our method uses orthonormal and overcomplete transformations that are unitary or redundant by design, avoiding feature selection bias.  
Moreover, unlike prior work, we introduce lightweight mappers that learn the geometry of each representation space to better align them during inference to enable effective linear probing.

\paragraph{Implicit bias from frequency}
Prior work has leveraged frequency information for representation learning in temporal sequences~\cite{tf_consistent, cost, unsupervised_periodicity}.
Some approaches, in line with ours, use the NT-Xent loss to maximize agreement between time-domain inputs and their frequency-transformed counterparts using the Fourier transform~\cite{tf_consistent} or spectrograms~\cite{multiformat, multimodal_ssl_audio}.
However, our method differs in two key ways.
First, these methods rely on additional encoders for each transformed view~\cite{tf_consistent, multiformat} and still require task-specific data augmentations.
In contrast, we replace augmentations entirely with principled transformations that generalize across datasets without task-specific tuning.
Second, previous work typically focuses on either global frequency features (via Fourier transform)~\cite{tf_consistent, temporal_freq_training} or localized frequency content (via spectrograms)~\cite{multiformat, multimodal_ssl_audio}.
Our method integrates both by jointly using orthonormal and overcomplete representations while remaining more efficient using lightweight latent space mappers instead of using separate modality-specific encoders.
\section{Conclusion}
In conclusion, we have shown that principled geometric transformations in the form of orthonormal bases and overcomplete frames are effective for self‑supervised representation learning.
By generating views through unitary and frame‐based projections, our method uses complementary manifolds without perturbing or enlarging the data.
Crucially, our approach achieves up to 15--20\% performance gains across nine datasets in five tasks, without relying on larger backbone architectures or handcrafted, domain-specific augmentations.
Our results underscore the importance of exploiting intrinsic geometric biases in data representations, opening a new avenue for SSL methods that prioritize mathematical structure over empirical trial and error.
We believe this work paves the way toward more generalizable, augmentation‑free self‑supervision across a wide range of domains.

\bibliography{main}
\bibliographystyle{unsrt}

\clearpage
\newpage
\section*{NeurIPS Paper Checklist}

\begin{enumerate}

\item {\bf Claims}
    \item[] Question: Do the main claims made in the abstract and introduction accurately reflect the paper's contributions and scope?
    \item[] Answer: \answerYes{} 
    \item[] Justification: We have provided the empirical results with extensive ablations to show the contributions.

\item {\bf Limitations}
    \item[] Question: Does the paper discuss the limitations of the work performed by the authors?
    \item[] Answer: \answerYes{} 
    \item[] Justification: Limitations are discussed in Section~\ref{sec:limitations}.

\item {\bf Theory assumptions and proofs}
    \item[] Question: For each theoretical result, does the paper provide the full set of assumptions and a complete (and correct) proof?
    \item[] Answer: \answerYes{} 
    \item[] Justification: Our theoretical results are given in Appendix~\ref{appen:proof}.

    \item {\bf Experimental result reproducibility}
    \item[] Question: Does the paper fully disclose all the information needed to reproduce the main experimental results of the paper to the extent that it affects the main claims and/or conclusions of the paper (regardless of whether the code and data are provided or not)?
    \item[] Answer: \answerYes{} 
    \item[] Justification: We have provided the implementation details in the main manuscript Section~\ref{ref:implementation} and Appendix~\ref{appendix:Implementation_Details}.

\item {\bf Open access to data and code}
    \item[] Question: Does the paper provide open access to the data and code, with sufficient instructions to faithfully reproduce the main experimental results, as described in supplemental material?
    \item[] Answer: \answerYes{}
    \item[] Justification: For double-blind review, we include our code in the supplementary materials.

\item {\bf Experimental setting/details}
    \item[] Question: Does the paper specify all the training and test details (e.g., data splits, hyperparameters, how they were chosen, type of optimizer, etc.) necessary to understand the results?
    \item[] Answer: \answerYes{} 
        \item[] Justification: We have provided details in Appendix~\ref{appendix:datasets} and~\ref{appendix:baselines_SSL}.

\item {\bf Experiment statistical significance}
    \item[] Question: Does the paper report error bars suitably and correctly defined or other appropriate information about the statistical significance of the experiments?
    \item[] Answer: \answerYes{} 
    \item[] Justification: We report the standard deviation across three runs with different random seeds to reflect the variability and statistical reliability of our results.

\item {\bf Experiments compute resources}
    \item[] Question: For each experiment, does the paper provide sufficient information on the computer resources (type of compute workers, memory, time of execution) needed to reproduce the experiments?
    \item[] Answer: \answerYes{} 
    \item[] Justification: We have provided the computer resources in Appendix~\ref{appendix:experiments}.
    
\item {\bf Code of ethics}
    \item[] Question: Does the research conducted in the paper conform, in every respect, with the NeurIPS Code of Ethics \url{https://neurips.cc/public/EthicsGuidelines}?
    \item[] Answer: \answerYes{} 
    \item[] Justification: The research was conducted in accordance with the NeurIPS Code of Ethics, and no ethical concerns or violations were identified.

\item {\bf Broader impacts}
    \item[] Question: Does the paper discuss both potential positive societal impacts and negative societal impacts of the work performed?
    \item[] Answer: \answerNA{} 
    \item[] Justification: This work proposes a general self-supervised learning framework for temporal data.
            It does not target a specific application area, and as such, it does not have direct or immediate societal impact.
    
\item {\bf Safeguards}
    \item[] Question: Does the paper describe safeguards that have been put in place for responsible release of data or models that have a high risk for misuse (e.g., pretrained language models, image generators, or scraped datasets)?
    \item[] Answer: \answerNA{} 
    \item[] Justification: Our paper does not include any pretrained language models, image generators or scraped datasets.

\item {\bf Licenses for existing assets}
    \item[] Question: Are the creators or original owners of assets (e.g., code, data, models), used in the paper, properly credited and are the license and terms of use explicitly mentioned and properly respected?
    \item[] Answer: \answerYes{} 
    \item[] Justification: We have added the relevant references in the appropriate sections of the manuscript.

\item {\bf New assets}
    \item[] Question: Are new assets introduced in the paper well documented and is the documentation provided alongside the assets?
    \item[] Answer: \answerYes{} 
    \item[] Justification: We release the source code accompanying our method as part of the supplementary material under a CC BY-NC-SA: Attribution-NonCommercial-ShareAlike - license. The code is documented and sufficient to reproduce the results presented in the paper. No new datasets or personally identifiable data are introduced.

\item {\bf Crowdsourcing and research with human subjects}
    \item[] Question: For crowdsourcing experiments and research with human subjects, does the paper include the full text of instructions given to participants and screenshots, if applicable, as well as details about compensation (if any)? 
    \item[] Answer: \answerNA{} 
    \item[] Justification: Our work does not involve any crowdsourcing or research with human subjects.

\item {\bf Institutional review board (IRB) approvals or equivalent for research with human subjects}
    \item[] Question: Does the paper describe potential risks incurred by study participants, whether such risks were disclosed to the subjects, and whether Institutional Review Board (IRB) approvals (or an equivalent approval/review based on the requirements of your country or institution) were obtained?
    \item[] Answer: \answerNA{} 
    \item[] Justification: Our work does not involve any crowdsourcing or research with human subjects.

\item {\bf Declaration of LLM usage}
    \item[] Question: Does the paper describe the usage of LLMs if it is an important, original, or non-standard component of the core methods in this research? Note that if the LLM is used only for writing, editing, or formatting purposes and does not impact the core methodology, scientific rigorousness, or originality of the research, declaration is not required.
    \item[] Answer: \answerNA{} 
    \item[] Justification: No large language models (LLMs) were used in the development or implementation of the core methods in this research. Any LLM use was limited to minor writing or editing support and did not impact the scientific contributions of the paper.

\end{enumerate}

\clearpage

\appendix
\begin{center}
    {\LARGE \bfseries Appendix}
\end{center}
\vspace{1em}

\section{Theoretical Analysis}
\label{appen:proof}
Here, we present complete proofs of our theoretical study, starting with notations.
We assume all the samples are absolutely summable, and finite.

\subsection{Representations and Notations}

\subsubsection{Orthonormal Bases and Overcomplete Frames}
\label{appen:notations_frames}
An orthonormal basis in a Hilbert space provides a complete set of mutually orthogonal unit-norm vectors. The Fourier transform forms such a basis for square-integrable signals over finite intervals.
For a discrete signal $\mathrm{x} $ of length $L$, the discrete Fourier transform (DFT) is defined as:

$$
\mathcal{F}_{\mathrm{x}}(k) = \frac{1}{\sqrt{L}} \sum_{n=0}^{L-1} x(n) \, e^{-j \frac{2\pi}{L} kn},
$$

where $k$ indexes the discrete frequencies. 
In this normalized form, the Fourier basis satisfies:

$$
\sum_k |\langle \mathrm{x}, e_k \rangle|^2 = \| \mathrm{x} \|^2,
$$

making it a tight frame (also known as a Parseval frame~\cite{finite_frames}).
A frame can be a generalization of a basis that allows for redundancy.
A set $\{\phi_i\} \subset \mathcal{H}$ is a frame for Hilbert space $\mathcal{H}$ if:

$$
A \| \mathrm{x} \|^2 \leq \sum_i |\langle \mathrm{x}, \phi_i \rangle|^2 \leq B \| \mathrm{x} \|^2,
$$

for all $x \in \mathcal{H}$ and constants $0 < A \leq B < \infty$.
Frames provide stability and flexibility, especially for non-stationary or noisy signals.

The Gabor wavelet transform is an example of such a redundant frame. It uses Gaussian-modulated sinusoids to achieve localized time-frequency decomposition, trading off time and frequency resolution optimally~\cite{Gabor_wavelet}.
The discrete Gabor wavelet transform for a signal \( \mathrm{x} \) is defined as in below.
\[
\mathcal{W}_{\mathrm{x}}(a, b) = \frac{1}{\sqrt{a}} \sum_{n=0}^{L-1} \mathrm{x} \, \psi\left( \frac{n-b}{a} \right),
\]
where \( a \) controls the scale and \( b \) controls the translation (time shift); we use 48 log-spaced scales and apply shifts at single time-step intervals.
The detailed implementations for the wavelet calculations are given in Appendix~\ref{appendix:Implementation_Details_transformations}.
The Gabor wavelet \( \psi(t) \) is defined as:
\[
\psi(t) = e^{-\frac{t^2}{2\sigma^2}} \, e^{j2\pi \xi t},
\]
where \( \sigma \) controls the width of the Gaussian envelope and \( \xi \) is the center frequency.
The Gabor wavelet captures localized oscillations and is well-suited for analyzing transient or non-stationary features in wide-range of signals.
In our work, the Fourier transform provides a global view of signal frequency content, while the Gabor wavelet transform enables fine-grained, localized analysis.
Combining these complementary perspectives improves the expressiveness of the learned representations.

The complete list of notations used throughout this manuscript is provided in Table~\ref{appen:notation_list}.

\subsubsection{Manifold}
In our method, we define each latent representation space as a manifold $\mathcal{M}$, similar to the manifold hypothesis~\cite{fefferman_testing_2016, hilbert_problem_manifold}, which states that high-dimensional data often lies on low-dimensional structures within the ambient space.
Since we generate three different latent representations (from time, Fourier, and wavelet domains), we consider a collection of manifolds, each corresponding to a specific transformation.
In our implementation, each manifold is assigned a fixed latent dimension of 128, resulting in a total latent dimension of 384 across the three manifolds.
For a fair comparison, we set the latent dimension to 384 for baseline methods, ensuring the linear classifier has the same number of parameters during fine-tuning.

\subsubsection{Notation List}
\label{appen:notation_list}
\begin{table}[h!]
    \centering
    \renewcommand{\arraystretch}{1.7}
    \begin{tabular}{ll}
        \textbf{Notation} & \textbf{Description} \\
        \hline
        $\boldsymbol{\mathrm{x}}$ & Temporal sequence represented as a bold lowercase symbol \\
        $\mathcal{F}_{\mathrm{x}}[k]$ & Fourier transformation of the temporal sequence with $k$ frequencies \\
        $\mathcal{W}_{\mathrm{x}}(a,b)$ & Gabor wavelet transformation of the temporal sequence \\
        $f_{\mathrm{x}}(.)$ & The encoder to obtain representations for the temporal sequence \\
        $f_{\mathcal{F}}(.)$ & The encoder to obtain representations for the Fourier transformed temporal sequence \\
        $f_{\mathcal{W}}(.)$ & The encoder to obtain representations for the wavelet transformed temporal sequence \\
        $g_{\mathrm{x}}(.)$ & The projector to obtain embeddings for the temporal sequence \\
        $g_{\mathcal{F} }(.)$ & The projector to obtain embeddings for the Fourier transform of the temporal sequence \\
        $g_{\mathcal{W} }(.)$ & The projector to obtain embeddings for the wavelet transformation of the temporal sequence \\
        $\boldsymbol{h}^{(t)}$ & The representations obtained from temporal sequence, i.e., $\boldsymbol{h}^{(t)} = f_{\mathrm{x}}(\mathrm{x})$ \\
        $\boldsymbol{h}^{(\mathcal{F})}$ & The representations obtained from the Fourier transformation of the temporal sequence \\
        $\boldsymbol{h}^{(\mathcal{W})}$ & The representations obtained from the Gabor wavelet transformation of the temporal sequence \\
        $\boldsymbol{z}^{(t)}$ & The embeddings of the sequence obtained from the projected representations, i.e. $\boldsymbol{z}^{(t)} = g_{t}(\boldsymbol{h}^{t})$ \\
        $\boldsymbol{z}^{(\mathcal{F})}$ & The embeddings of the Fourier transformed temporal sequence \\
        $\boldsymbol{z}^{(\mathcal{W})}$ & The embeddings of the wavelet transformed temporal sequence \\
        $\Phi^{t \rightarrow \mathcal{F}}_{\boldsymbol{h}}$ & The representation mapping function from time to Fourier domain \\
        $\Phi^{t \rightarrow \mathcal{W}}_{\boldsymbol{h}}$ & The representation mapping function from time to Wavelet domain \\
        $\Phi^{t \rightarrow \mathcal{F}}_{\boldsymbol{z}}$ & The embedding mapping function from time to Fourier domain \\
        $\Phi^{t \rightarrow \mathcal{W}}_{\boldsymbol{z}}$ & The embedding mapping function from time to Wavelet domain \\
        $\langle \boldsymbol{a}, \boldsymbol{b} \rangle $ & The inner product of two vectors $\boldsymbol{a}$ and $\boldsymbol{b}$ \\
        $\mathrm{sim}(\boldsymbol{z}_i^{(d)}, \boldsymbol{z}_j^{(d')})$ & Cosine similarity between the embeddings \\
        ${\perp}$ & Perpendicular \\ 
        $\tau$ & Temperature coefficient for NT-Xent loss \\
        $\mathcal{M}$ & Manifold notation \\
        $\mathcal{L}$ & A loss function, i.e., cross-entropy. \\
        \bottomrule
    \end{tabular}
    \vspace{3mm}
    \caption{Detailed list of notations used in this work}
    \label{tab:notations}
\end{table}

\clearpage

\subsection{Proof for Proposition~\ref{prop:angle_pairwise}}
\begin{proposition}[Angle Concentration vs. Pairwise Spread]\label{append_prop:angle_pairwise}
Let \( \boldsymbol{h}^{(t)}, \boldsymbol{h}^{(\mathcal{F})} \sim \mathrm{Unif}(S^{d-1}) \), where \( \boldsymbol{h}^{(\mathcal{F})} = f_{\mathcal{F}}(\mathcal{F}(\mathrm{x})) \).  
Although individual samples across latent spaces tend toward orthogonality, the pairwise angular difference \( \Delta_{ij} \) between distinct samples can span the full range up to \( \pi \).
\begin{gather}
\arccos\big( \langle \boldsymbol{h}^{(t)}, \boldsymbol{h}^{(\mathcal{F})} \rangle \big) = \frac{\pi}{2}, \hspace{2mm} \text{while} \hspace{2mm}
\arccos (\langle \boldsymbol{h}_i^{(t)}, \boldsymbol{h}_j^{(t)} \rangle ) - \arccos (\langle \boldsymbol{h}_i^{(\mathcal{F})}, \boldsymbol{h}_j^{(\mathcal{F})} \rangle ) 
= \Delta_{ij} \leq \pi
\end{gather}
\end{proposition}
\begin{proof}
The first part follows the Chernoff on the spherical cap for angle concentration,
Let  
\(\boldsymbol h^{(t)},\boldsymbol h^{(\mathcal F)}\stackrel{\text{i.i.d.}}{\sim}\mathrm{Unif}(S^{d-1})\)  
and write \(\xi:=\langle\boldsymbol h^{(t)},\boldsymbol h^{(\mathcal F)}\rangle\).

\begin{equation}
\mathbb E[\xi]=0,\qquad
\operatorname{Var}[\xi]=\frac1d .
\end{equation}

Lévy’s concentration~\cite{chavel_eigenvalues_1984, chavel_isoperimetric_2001} gives, for every
\(\varepsilon\in(0,1)\),

\begin{equation}
\Pr \bigl(|\xi|>\varepsilon\bigr)
\;\le\;2\exp \Bigl(-\tfrac{(d-2)\varepsilon^{2}}{2}\Bigr)
\;\xrightarrow[d\to\infty]{}0 
\end{equation}

Hence \(\xi\xrightarrow{p}0\) and, by continuity of \(\arccos\) at \(0\),

\begin{equation}
\arccos\xi\xrightarrow{p}\frac{\pi}{2}.
\end{equation}

Let distinct indices \(i\neq j\) while setting,

\begin{equation}
V:=\boldsymbol h_i^{(t)\!\perp}\cap\boldsymbol h_j^{(t)\!\perp},
\qquad
\dim V=d-2\ge1\;\;(d\ge3).
\end{equation}

Choose orthonormal \(u,u_{\perp}\in V\).  
For any \(\phi\in[0,\pi]\) define

\begin{equation}
\boldsymbol h_i^{(\mathcal F)}:=u,\qquad
\boldsymbol h_j^{(\mathcal F)}:=\cos\phi\,u+\sin\phi\,u_{\perp}
\end{equation}

Then

\begin{equation}
\langle\boldsymbol h^{(t)}_i,\boldsymbol h^{(\mathcal F)}_i\rangle
=\langle\boldsymbol h^{(t)}_j,\boldsymbol h^{(\mathcal F)}_j\rangle
=0,
\qquad
\langle\boldsymbol h^{(\mathcal F)}_i,\boldsymbol h^{(\mathcal F)}_j\rangle
=\cos\phi,
\end{equation}

Let

\begin{equation}
\theta_{ij}^{(t)}:=\arccos\langle\boldsymbol h^{(t)}_i,\boldsymbol h^{(t)}_j\rangle,
\qquad
\theta_{ij}^{(\mathcal F)}:=\arccos\langle\boldsymbol h^{(\mathcal F)}_i,
                                          \boldsymbol h^{(\mathcal F)}_j\rangle
=\phi.
\end{equation}

Thus

\begin{equation}
\Delta_{ij}:=\theta_{ij}^{(t)}-\theta_{ij}^{(\mathcal F)}
=\theta_{ij}^{(t)}-\phi,
\end{equation}

and varying \(\phi\) over \([0,\pi]\) makes \(\Delta_{ij}\) sweep the full
interval \([\theta_{ij}^{(t)}-\pi,\;\theta_{ij}^{(t)}]\subset[-\pi,\pi]\).
In particular, \(\phi=0\) or \(\pi\) yields \(|\Delta_{ij}|=\pi\).
\end{proof}
Therefore, even though the angles from same samples concentrate at \(\pi/2\), pairwise discrepancies can reach any value up to the maximal \(\pi\).

In practice, although $\boldsymbol{h}^{(t)}$ and $\boldsymbol{h}^{(\mathcal F)}$ are coupled via the same loss, we conjecture that the high-dimensional geometry and symmetric repulsion of negatives make their joint distribution approximate to the independent uniforms on $S^{d-1}$~\cite{alignment_uniformity}.
This justifies modeling both latent spaces as samples from $\mathrm{Unif}(S^{d-1})$ when analyzing angular gaps between non-matched pairs.

\clearpage

\subsection{Proof for Proposition~\ref{prop:no_invariance_collapse}}
\begin{proposition}
\label{prop_appendix:no_invariance_collapse}
Let $f_d^{\ast}$ denote an optimal encoder under NT-Xent for domain $d \in \{t,\mathcal{F},\mathcal{W}\}$.  
If for some unintended transformation $W$ the encoder is invariant, i.e., $f_d^{\ast}(W\mathrm{x}) = f_d^{\ast}(\mathrm{x}),$ then for any anchor sample $x$ the NT-Xent loss across domains is lower bounded by the number of negatives.
\[
    \ell\!\left(\boldsymbol{z}_i^{(d)}, \boldsymbol{z}_j^{(d')}\right) \;\;\geq\;\; \log(K+1) > 0,
\]
where $K \geq 1$ is the number of negatives that become near-positives due to the invariance.
\end{proposition}

\begin{proof}
For each domain $d \in \{t,\mathcal{F},\mathcal{W}\}$, let $f_d:\mathcal{X}\!\to\!\mathbb{R}^m$ be the encoder and $g_d$ the projection head. We write the (unit-normalized) embedding as
\begin{equation}
\boldsymbol{z}^{(d)}(\mathrm{x}) \coloneqq \frac{g_d(f_d(\mathrm{x}))}{\lVert g_d(f_d(\mathrm{x}))\rVert_2}\in\mathbb{S}^{m-1}, 
\quad 
\text{and }~\mathrm{sim}(u,v)\coloneqq u^\top v \in [-1,1]
\end{equation}
For an anchor $\mathrm{x}$ in domain $d$ and its positive view in domain $d'\!\neq\! d$ (same underlying sample), the NT-Xent loss is
\begin{equation}
\label{eq:ntxent-ddprime}
\ell_{d,d'}(\mathrm{x})
= -\log \frac{\exp\big(\mathrm{sim}(\boldsymbol{z}^{(d)}(\mathrm{x}), \boldsymbol{z}^{(d')}(\mathrm{x}))/\tau\big)}
{\exp\big(\mathrm{sim}(\boldsymbol{z}^{(d)}(\mathrm{x}), \boldsymbol{z}^{(d')}(\mathrm{x}))/\tau\big) + \sum\limits_{k\neq i}\exp\big(\mathrm{sim}(\boldsymbol{z}^{(d)}(\mathrm{x}), \boldsymbol{z}^{(d')}(\mathrm{x}_k))/\tau\big)}\,,
\end{equation}
with temperature $\tau>0$. The full instance-discrimination objective $\mathcal{L}_{\mathrm{ID}}$ (Equation~\ref{eq:ID}) averages/sums $\ell_{d,d'}$ over all ordered domain pairs and batch samples.

Suppose there exists an unintended transformation $W$ and a domain $d \in \{t,\mathcal{F},\mathcal{W}\}$ such that
\begin{equation}
\label{eq:invar}
f_d^\ast(Wx)=f_d^\ast(\mathrm{x})
\end{equation}
In a batch, let $S\subset\{k\neq i\}$ be indices of \emph{near-positive} negatives created by the invariance:
\begin{equation}
\label{eq:nearpos}
\mathrm{sim} \big(\boldsymbol{z}^{(d)}(\mathrm{x}),\, \boldsymbol{z}^{(d')}(\mathrm{x}_k)\big) \ge 1-\delta
\quad \text{for all } k\in S,
\end{equation}
with $|S|=K\ge 1$ and some $\delta\in[0,1)$. In the exact invariance case, $\delta=0$.

\paragraph{Per-pair lower bound.}
Fix $(d,d')$ and an anchor sample $\mathrm{x}$.
We can define the positive similarity between embeddings as $s_{\mathrm{pos}}\coloneqq\!\mathrm{sim}(\boldsymbol{z}^{(d)},\boldsymbol{z}^{(d')})$.
If we split the negative set into $S$ (the $K$ near-positives) and $R$ (the rest).
From \eqref{eq:ntxent-ddprime} and \eqref{eq:nearpos}, we can write the loss function as in Equation~\ref{eq:loss_app}
\begin{equation}\label{eq:loss_app}
\ell_{d,d'}(\mathrm{x})
= -\log \frac{e^{s_{\mathrm{pos}}/\tau}}
{e^{s_{\mathrm{pos}}/\tau} + \sum_{k\in S} e^{\mathrm{sim}(\boldsymbol{z}^{(d)}(\mathrm{x}), \boldsymbol{z}^{(d')}(\mathrm{x}_k))/\tau} + \sum_{r\in R} e^{\mathrm{sim}(\cdot)/\tau}}
\;\;\ge\;\;
-\log \frac{e^{s_{\mathrm{pos}}/\tau}}
{e^{s_{\mathrm{pos}}/\tau} + K\, e^{(1-\delta)/\tau}}
\end{equation}
Factor $e^{s_{\mathrm{pos}}/\tau}$ from the denominator,
\begin{align*}
\ell_{d,d'}(\mathrm{x})
&\ge -\log \frac{1}{1 + K\, e^{\frac{1-\delta-s_{\mathrm{pos}}}{\tau}}}
= \log\!\Big(1 + K\, e^{\frac{1-\delta-s_{\mathrm{pos}}}{\tau}}\Big)
\end{align*}
At optimum, $s_{\mathrm{pos}}=1$ (or $s_{\mathrm{pos}}\ge 1-\varepsilon$ in the approximate case). Since the RHS is decreasing in $s_{\mathrm{pos}}$, the weakest bound (i.e., smallest lower bound that still holds) is obtained at $s_{\mathrm{pos}}=1$, giving
\begin{equation}
\label{eq:keybound}
\ell_{d,d'}(\mathrm{x}) \;\ge\; \log\!\Big(1 + K\, e^{-\delta/\tau}\Big)
\end{equation}
In particular, for exact near-positives ($\delta=0$),
\(
\ell_{d,d'}(\mathrm{x})\ge \log(K+1).
\)

As is standard, we assume that at the NT-Xent optimum positive pairs align~\cite{chaos_is_a_ladder, alignment_uniformity}, i.e., $\mathrm{sim}(\boldsymbol{z}^{(d)}(\mathrm{x}), \boldsymbol{z}^{(d')}(\mathrm{x}))\!=\!1$.
To extend this proof, we have also provided a case which also covers the approximate case.  
If $s_{\mathrm{pos}}\ge 1-\varepsilon$ for some small $\varepsilon\ge 0$, then
\[
\ell_{d,d'}(x)
\;\ge\; \log\!\Big(1 + K\, e^{\frac{\varepsilon-\delta}{\tau}}\Big)
\;\ge\; \log\!\Big(1 + K\, e^{-\delta/\tau}\Big),
\]
because $\varepsilon\ge 0$.
Thus, Equation~\ref{eq:keybound} still holds.
\end{proof}

The bound in Equation~\ref{eq:keybound} is tight when $s_{\mathrm{pos}}=1$ and exactly $K$ negatives have similarity $1-\delta$ while all others contribute negligibly, so the denominator equals $e^{1/\tau} + K e^{(1-\delta)/\tau}$.

Extending to the full objective, recall that $\mathcal{L}_{\mathrm{ID}}$ (Eq.~\ref{eq:ID}) sums/averages $\ell_{d,d'}$ over all ordered domain pairs $(d,d')$ with $d\neq d'$. For a fixed anchor $\mathrm{x}$, if invariance induces $K_{d,d'}$ near-positive negatives with parameter $\delta_{d,d'}$ for pair $(d,d')$, then applying Equation~\ref{eq:keybound} to each pair and summing gives
\begin{equation}
\sum_{d\neq d'} \ell_{d,d'}(\mathrm{x})
\;\ge\;
\sum_{d\neq d'} \log\!\Big(1 + K_{d,d'}\, e^{-\delta_{d,d'}/\tau}\Big),
\end{equation}
and by linearity of expectation,
\begin{equation}
\mathcal{L}_{\mathrm{ID}}
\;\ge\;
\mathbb{E}_{\mathrm{x}}\!\left[\sum_{d\neq d'} \log\!\Big(1 + K_{d,d'}\, e^{-\delta_{d,d'}/\tau}\Big)\right]
\end{equation}
Asymptotically, for fixed $\tau$ and bounded $\delta_{d,d'}$, each affected pair contributes $\Theta(\log(1+K_{d,d'}))=\Omega(\log K_{d,d'})$. If a nonzero fraction $p$ of a batch of size $B$ yields near-positives per affected pair ($K_{d,d'}=\Theta(B)$), then $\ell_{d,d'}(\mathrm{x})=\Omega(\log B)$ and consequently $\mathcal{L}_{\mathrm{ID}}=\Omega \Big(\sum_{d\neq d'} \log B\Big)$ over those pairs. Thus any unintended invariance that produces even a single near-positive per ordered pair enforces a nontrivial lower bound; if collisions scale with batch size, the bound grows at least logarithmically in $B$.

Unintended invariances therefore inflate the NT-Xent denominator through near-positive negatives, yielding the lower bound in Equation~\ref{eq:keybound}.
We complete the proof by showing the objective cannot be minimized without bound when such invariances hold.

\clearpage

\section{Algorithm}
\label{appen:Algorithm}
In this section, we present the pseudocode for our method during pre-training and inference.
Algorithm~\ref{alg:method} describes the training procedure, which takes a sample, $\mathrm{x}$, and model components as inputs, and outputs the trained encoder and latent space mappers.

\begin{algorithm}
\caption{\label{alg:method} Pre-training algorithm for the proposed method}
\begin{algorithmic}[1]
\setlength{\itemsep}{1.4ex} 
\State \textbf{Input:} $\mathrm{x}$, $\mathcal{F}_{\mathrm{x}}$, $\mathcal{W}_{\mathrm{x}}$, and the required models, i.e., $f_{\mathrm{x}}(.)$.  
\State \textbf{Output:} $f_{\mathrm{x}}(.)$, $\Phi^{t \rightarrow \mathcal{F}}_{\boldsymbol{h}}$, $\Phi^{t \rightarrow \mathcal{W}}_{\boldsymbol{h}}$
\Comment{\small The output of the pre-training is the single encoder with mappers}
\State $\boldsymbol{h}^{(\mathrm{t})} = f_{\mathrm{x}}(\mathrm{x})$, $\boldsymbol{h}^{(\mathcal{F})} = f_{\mathcal{F}}(\mathcal{F_{\mathrm{x}}})$, $\mathbf{h}^{(\mathcal{W})} = f_{\mathcal{W}}(\mathcal{W}_{\mathrm{x}})$
\Comment{\small Obtain representations for each input}
\State $\boldsymbol{z}^{(t)} = g_{\mathrm{x}}(\boldsymbol{h}^{(t)})$, $\boldsymbol{z}^{(\mathcal{F})} = g_{\mathcal{F}}( \boldsymbol{h}^{(\mathcal{F})} )$, $\boldsymbol{z}^{(\mathcal{W})} = g_{\mathcal{W}}(\boldsymbol{h}^{(\mathcal{W})})$
\Comment{\small Obtain embeddings for each representation}
\State $\mathcal{L}_{\mathrm{ID}}=\sum_{\substack{d,d'\in\{t,\mathcal{F},\mathcal{W}\}\\d\neq d'}} \frac{1}{2N}\sum_{k=1}^N
\Bigl[
   \ell\bigl(\boldsymbol{z}_{k-1}^{(d)}, \boldsymbol{z}_{k}^{(d')}\bigr)
 + \ell\bigl(\boldsymbol{z}_{k}^{(d)}, \boldsymbol{z}_{k-1}^{(d')}\bigr)
\Bigr]$
\Comment{\small Instance discrimination loss}
\State $\boldsymbol{z}^{(\mathcal{F})}_{\text{est}} = \Phi^{t\rightarrow\mathcal{F}}_{\boldsymbol{z}}(\boldsymbol{z}^{(t)})$, $\boldsymbol{z}^{(\mathcal{W})}_{\text{est}} = \Phi^{t\rightarrow\mathcal{W}}_{\boldsymbol{z}}(\boldsymbol{z}^{(t)})$
\Comment{\small Obtain the estimated embeddings for both transformations}
\State $\mathcal{L}_{\text{map}} = 
\frac{1}{N} \left\| \Phi_{\boldsymbol{z}}^{t \rightarrow \mathcal{F}} \left( \boldsymbol{z}^{(t)} \right) - \boldsymbol{z}^{(\mathcal{F})} \right\|_1 
+ 
\frac{1}{N} \left\| \Phi_{\boldsymbol{z}}^{t \rightarrow \mathcal{W}} \left(\boldsymbol{z}^{(t)} \right) - \boldsymbol{z}^{(\mathcal{W})} \right\|_1$
\Comment{\small Mapping loss for embeddings}
\Statex\rule{\linewidth}{0.4pt}
\Statex \textbf{Freeze} $\{f_{\mathrm{x}},\,f_{\mathcal{F}}, \, f_{\mathcal{W}}\}$; \textbf{Omit}, $\{g_{\mathrm{x}},\, g_{\mathcal{W}},\,g_{\mathcal{W}},\, \Phi^{t \rightarrow \mathcal{F}}_{\boldsymbol{z}}, \Phi^{t \rightarrow \mathcal{M}}_{\boldsymbol{z}} \}$; \textbf{Train} $\{\Phi_{\boldsymbol{h}}^{t\to\mathcal F},\Phi_{\boldsymbol{h}}^{t\to\mathcal W}\}$
\Statex\rule{\linewidth}{0.4pt}
\State $\boldsymbol{h}^{(\mathrm{t})} = f_{\mathrm{x}}(\mathrm{x})$, $\boldsymbol{h}^{(\mathcal{F})} = f_{\mathcal{F}}(\mathcal{F_{\mathrm{x}}})$, $\mathbf{h}^{(\mathcal{W})} = f_{\mathcal{W}}(\mathcal{W}_{\mathrm{x}})$
\Comment{\small Obtain representations using trained models}
\State $\boldsymbol{h}^{(\mathcal{F})}_{\text{est}} = \Phi^{t\rightarrow\mathcal{F}}_{\boldsymbol{h}}(\boldsymbol{h}^{(t)})$, $\boldsymbol{h}^{(\mathcal{W})}_{\text{est}} = \Phi^{t\rightarrow\mathcal{W}}_{\boldsymbol{h}}(\boldsymbol{h}^{(t)})$
\Comment{\small Obtain the estimated representations for transformations}
\State $\mathcal{L}_{\text{map}} = 
\frac{1}{N} \left\| \Phi_{\boldsymbol{h}}^{t \rightarrow \mathcal{F}} \left( \boldsymbol{h}^{(t)} \right) - \boldsymbol{h}^{(\mathcal{F})} \right\|_1 
+ 
\frac{1}{N} \left\| \Phi_{\boldsymbol{h}}^{t \rightarrow \mathcal{W}} \left(\boldsymbol{h}^{(t)} \right) - \boldsymbol{h}^{(\mathcal{W})} \right\|_1$
\Comment{\small Mapping loss for representations}
\State \textbf{Return:} $f_{\mathrm{x}}(.)$, $\Phi^{t \rightarrow \mathcal{F}}_{\boldsymbol{h}}$, $\Phi^{t \rightarrow \mathcal{W}}_{\boldsymbol{h}}$
\end{algorithmic}
\end{algorithm}

Algorithm~\ref{alg:method_inference} outlines the inference process, using only the main encoder and lightweight mappers.
After applying the mappers, we concatenate the resulting representations for linear probing.
During linear probing, both the main encoder and the mappers are kept frozen.

\begin{algorithm}
\caption{\label{alg:method_inference} The proposed method for inference}
\begin{algorithmic}[1]
\setlength{\itemsep}{1.4ex} 
\State \textbf{Input:} $\mathrm{x}$, $f_{\mathrm{x}}(.)$, $\Phi^{t \rightarrow \mathcal{F}}_{\boldsymbol{h}}$, and $\Phi^{t \rightarrow \mathcal{W}}_{\boldsymbol{h}}$.  
\State \textbf{Output:} $\boldsymbol{h}$
\Comment{\small The output of the pre-training is the representations}
\State $\boldsymbol{h}^{(\mathrm{t})} = f_{\mathrm{x}}(\mathrm{x})$
\Comment{\small Obtain representations for inputs}
\State $\boldsymbol{h}_{\text{est}}^{(\mathcal{F})} = \Phi^{t \rightarrow \mathcal{F}}_{\boldsymbol{h}}(\boldsymbol{h}^{(t)})$, \, $\boldsymbol{h}_{\text{est}}^{(\mathcal{W})} = \Phi^{t \rightarrow \mathcal{W}}_{\boldsymbol{h}}(\boldsymbol{h}^{(t)})$ 
\Comment{\small Obtain the estimated representations in other domains}
\State $ \boldsymbol{h}
  = \bigl[\,
      \boldsymbol{h}^{(t)} \;;\;
      \boldsymbol{h}_{\mathrm{est}}^{(\mathcal{F})} \;;\;
      \boldsymbol{h}_{\mathrm{est}}^{(\mathcal{W})}
    \bigr]$
\Comment{\small Concatenate features from all domains}
\State \textbf{Return:} $\boldsymbol{h}$
\end{algorithmic}
\end{algorithm}

\clearpage

\section{Pairwise Distances}
\label{appendix:pair_wise_distances}
\begin{figure}[h]
    \centering

    \begin{subfigure}[t]{0.45\textwidth}
        \centering
        \includegraphics[width=\linewidth]{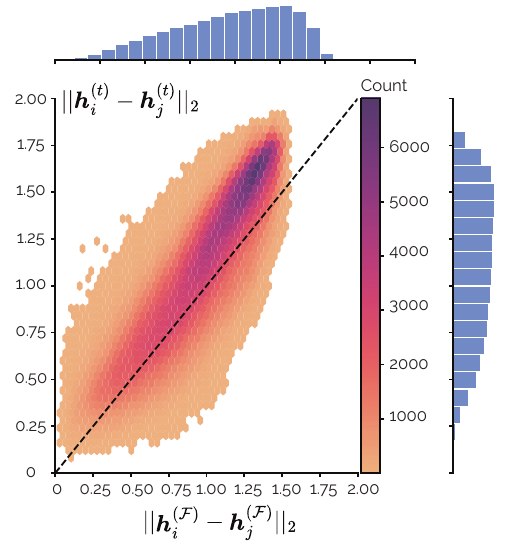}
        \caption{Pairwise $\ell_2$ distance comparison between time and Fourier domain latent spaces on IEEE SPC12}
    \end{subfigure}
    \hfill
    \begin{subfigure}[t]{0.45\textwidth}
        \centering
        \includegraphics[width=\linewidth]{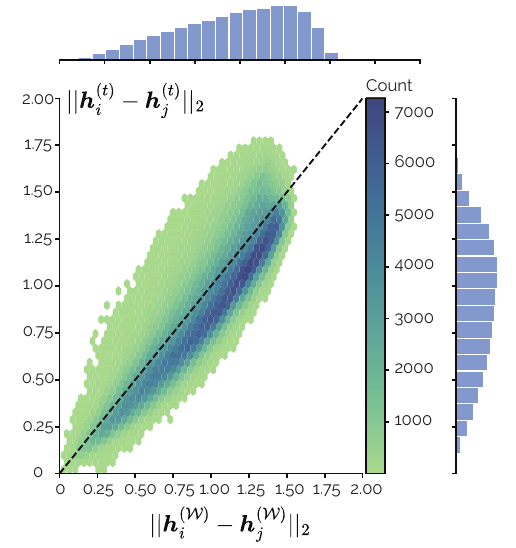}
        \caption{Pairwise $\ell_2$ distance comparison between time and wavelet domain latent spaces on IEEE SPC12}
    \end{subfigure}

    \vspace{2mm}

    \begin{subfigure}[t]{0.45\textwidth}
        \centering
        \includegraphics[width=\linewidth]{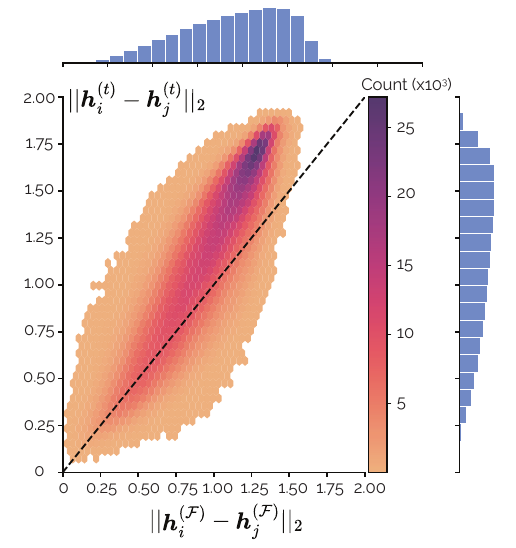}
        \caption{Pairwise $\ell_2$ distance comparison between time and Fourier domain latent spaces on IEEE SPC22}
    \end{subfigure}
    \hfill
    \begin{subfigure}[t]{0.45\textwidth}
        \centering
        \includegraphics[width=\linewidth]{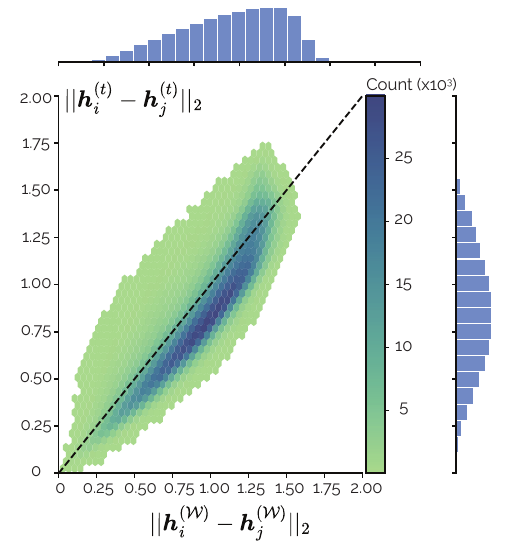}
        \caption{Pairwise $\ell_2$ distance comparison between time and wavelet domain latent spaces on IEEE SPC22}
    \end{subfigure}
    \caption{Pairwise $\ell_2$ distance comparisons across domains and datasets for \textit{heart rate} estimation.\label{appendix_l2_ppg}}
\end{figure}

\begin{figure}[h]
    \centering

    \begin{subfigure}[t]{0.45\textwidth}
        \centering
        \includegraphics[width=\linewidth]{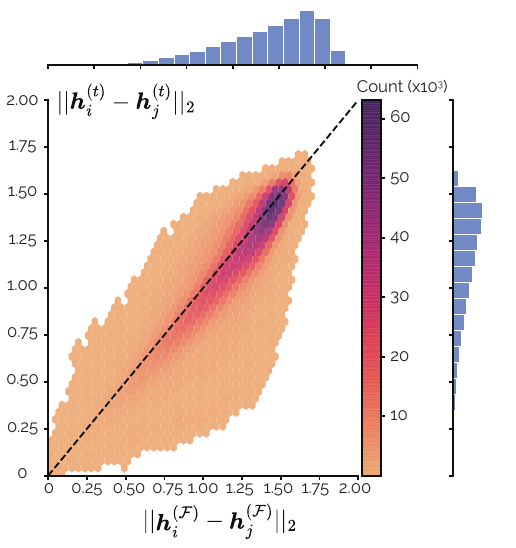}
        \caption{Pairwise $\ell_2$ distance comparison between time and Fourier domain latent spaces on HHAR}
    \end{subfigure}
    \hfill
    \begin{subfigure}[t]{0.45\textwidth}
        \centering
        \includegraphics[width=\linewidth]{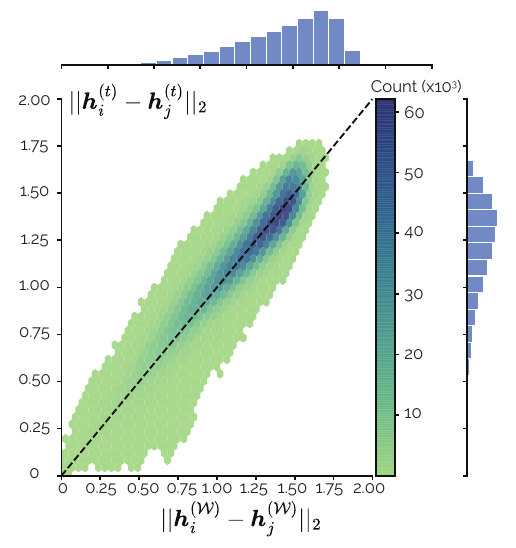}
        \caption{Pairwise $\ell_2$ distance comparison between time and wavelet domain latent spaces on HHAR}
    \end{subfigure}

    \vspace{2mm}

    \begin{subfigure}[t]{0.45\textwidth}
        \centering
        \includegraphics[width=\linewidth]{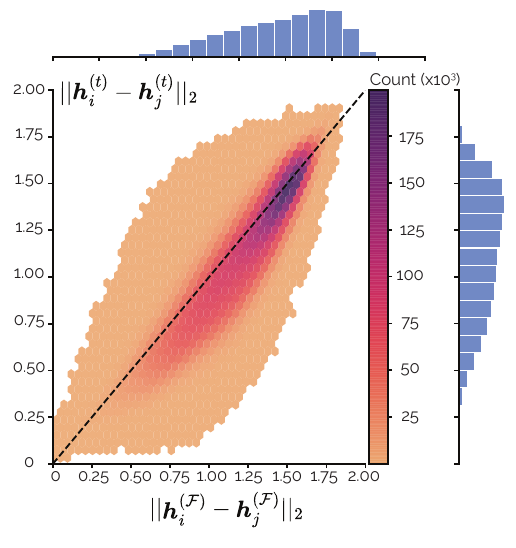}
        \caption{Pairwise $\ell_2$ distance comparison between time and Fourier domain latent spaces on USC}
    \end{subfigure}
    \hfill
    \begin{subfigure}[t]{0.45\textwidth}
        \centering
        \includegraphics[width=\linewidth]{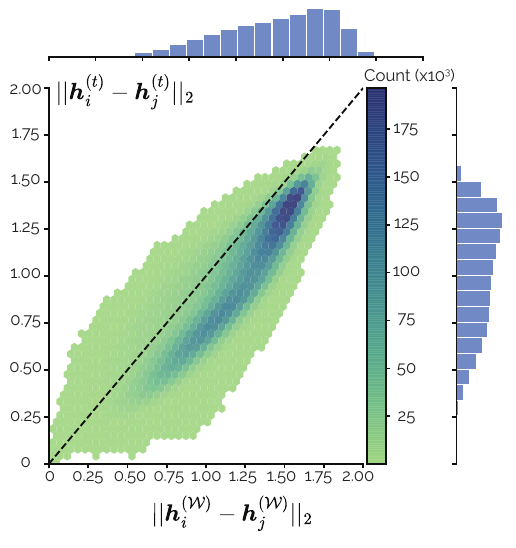}
        \caption{Pairwise $\ell_2$ distance comparison between time and wavelet domain latent spaces on USC}
    \end{subfigure}
    
    \caption{Pairwise $\ell_2$ distance comparisons across domains and datasets for \textit{activity} recognition.\label{appendix_l2_imu}}
    
\end{figure}

\begin{figure}[h]
    \centering

    \begin{subfigure}[t]{0.45\textwidth}
        \centering
        \includegraphics[width=\linewidth]{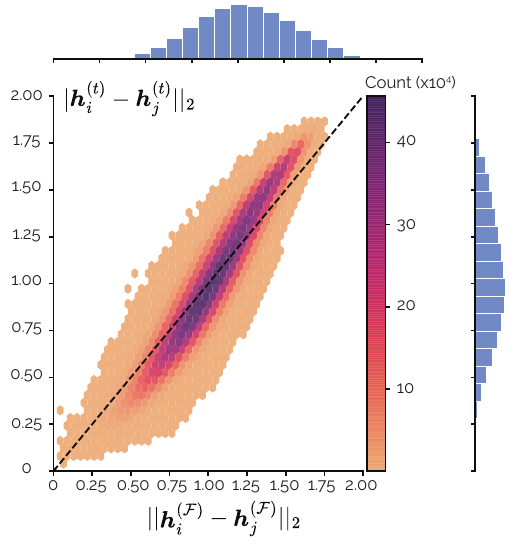}
        \caption{Pairwise $\ell_2$ distance comparison between time and Fourier domain latent spaces on CPSC}
    \end{subfigure}
    \hfill
    \begin{subfigure}[t]{0.45\textwidth}
        \centering
        \includegraphics[width=\linewidth]{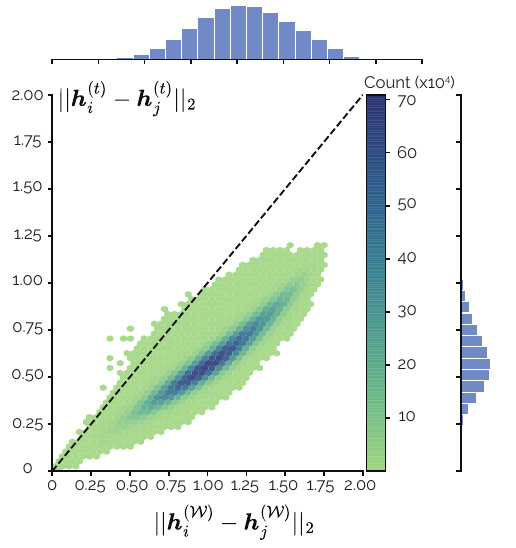}
        \caption{Pairwise $\ell_2$ distance comparison between time and wavelet domain latent spaces on CPSC}
    \end{subfigure}

    \vspace{2mm}

    \begin{subfigure}[t]{0.45\textwidth}
        \centering
        \includegraphics[width=\linewidth]{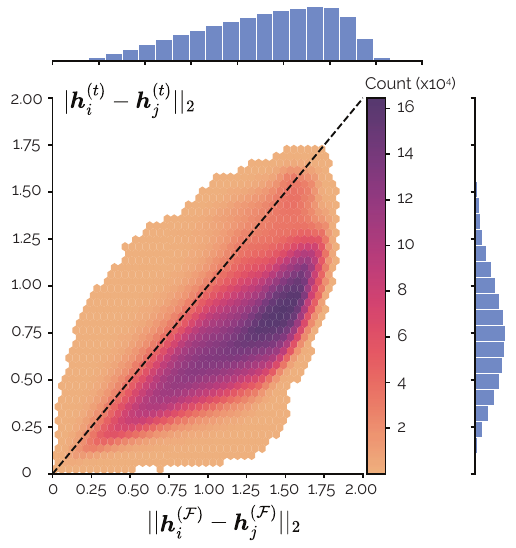}
        \caption{Pairwise $\ell_2$ distance comparison between time and Fourier domain latent spaces on Sleep}
    \end{subfigure}
    \hfill
    \begin{subfigure}[t]{0.45\textwidth}
        \centering
        \includegraphics[width=\linewidth]{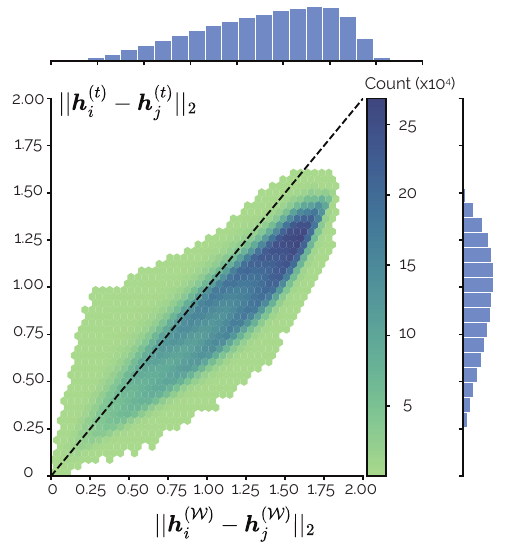}
        \caption{Pairwise $\ell_2$ distance comparison between time and wavelet domain latent spaces on Sleep}
    \end{subfigure}
    
    \caption{Pairwise $\ell_2$ distance comparisons across domains and datasets for \textit{cardiovascular disease} and \textit{sleep} classification.\label{appendix_l2_cpsc_sleep}}
    
\end{figure}

In addition to angular comparisons, we report the $\ell_2$ distances between sample pairs in the time, Fourier, and wavelet domains.
Figures~\ref{appendix_l2_ppg},~\ref{appendix_l2_imu} and~\ref{appendix_l2_cpsc_sleep} visualize these results.
If pairwise distances were preserved across latent spaces, all points would lie along the $y = x$ line---i.e., the distance between samples $i$ and $j$ in the time-domain latent space $\boldsymbol{h}^{(t)}$ would match that in the transformed domain $\boldsymbol{h}^{(d)}$, where \( d \in \{\mathcal{F}, \mathcal{W}\} \).


\clearpage

We conducted this investigation across tasks involving both single- and multi-channel temporal data. 
It is worth noting that while some datasets exhibit closer alignment between latent spaces, we observed consistent and non-negligible deviations across all tasks, indicating that latent space geometries differ across applications.

\clearpage

These figures reveal substantial deviations from $y=x$ line, indicating that distances between samples ($i \neq j$) in the latent space vary across domains.
Importantly, these discrepancies are consistent despite all encoders being trained jointly with the same objective.
These observations further support our motivation for leveraging multiple latent spaces to capture complementary structure in the data.

\clearpage

\section{Additional Experiments}
\label{appendix:additional_experiments}

\subsection{Embedding mappers}
We have employed embedding space mappers ($\Phi_{\boldsymbol{z}}^{t \rightarrow d}$ ) after projection layers to improve predictability across latent spaces and reduce estimation error.
In this section, we have presented the ablation experiments regarding the performance when the embedding mappers are excluded (w/o $\Phi_{\boldsymbol{z}}^{t \rightarrow d}$) from our method.
The results are given in Tables~\ref{tab:performance_ppg_ablation_2},~\ref{tab:performance_activity_ablation_2} and~\ref{tab:performance_ecg_eeg_ablation_2}.

\begin{table*}[h]
\centering
\caption{Further ablation on proposed method in \textit{PPG} datasets for HR estimation}
\begin{adjustbox}{width=\columnwidth,center}
\label{tab:performance_ppg_ablation_2}
\renewcommand{\arraystretch}{1.1}
\begin{tabular}{@{}lllllllllll@{}}
\toprule
\multirow{2}{*}{Method} & \multicolumn{3}{l}{IEEE SPC12} & \multicolumn{3}{l}{IEEE SPC22} & \multicolumn{3}{l}{DaLiA$_{PPG}$} \\ 
\cmidrule(r{15pt}){2-4}  \cmidrule(r{15pt}){5-7}  \cmidrule(r{15pt}){8-10} 
& MAE $\downarrow$ & RMSE $\downarrow$ & $\rho$ $\uparrow$ & MAE $\downarrow$ & RMSE $\downarrow$ & $\rho$ $\uparrow$ & MAE $\downarrow$ & RMSE $\downarrow$ & $\rho$ $\uparrow$ \\
\midrule
Ours & \textbf{8.84}\small$\pm$0.50 & \textbf{14.37}\small$\pm$0.95 & \textbf{82.67}\small$\pm$1.30 & \textbf{14.06}\small$\pm$1.09 & \textbf{21.48}\small$\pm$2.01 & \textbf{54.88}\small$\pm$1.89 & 9.13\small$\pm$0.20 & 16.92\small$\pm$0.56 & 63.72\small$\pm$0.06 & \\
w/o $\Phi_{\boldsymbol{z}}^{t \rightarrow d}$ & 9.87 \small(\textcolor{WildStrawberry}{+1.03})  & 15.85 \small(\textcolor{WildStrawberry}{+1.48}) & 79.78 \small(\textcolor{WildStrawberry}{-2.89}) & 15.17 \small(\textcolor{WildStrawberry}{+1.11}) & 24.75 \small(\textcolor{WildStrawberry}{+2.27}) & 53.19 \small(\textcolor{WildStrawberry}{-1.69}) & 9.83 \small(\textcolor{WildStrawberry}{+0.70}) & 17.68 \small(\textcolor{WildStrawberry}{+0.76}) & 61.15 \small(\textcolor{WildStrawberry}{-2.57})  \\

\noalign{\vskip 1mm}
\hline  
\noalign{\vskip 1mm}

Lin. $\Phi_{\boldsymbol{h}}^{t \rightarrow d}$ & 9.77 \small(\textcolor{WildStrawberry}{+0.93})  & 16.27 \small(\textcolor{WildStrawberry}{+1.90}) & 78.77 \small(\textcolor{WildStrawberry}{-3.90}) & 15.61 \small(\textcolor{WildStrawberry}{+1.55}) & 24.10 \small(\textcolor{WildStrawberry}{+2.58}) & 54.10 \small(\textcolor{WildStrawberry}{-0.78}) & 9.23 \small(\textcolor{WildStrawberry}{+0.10}) & 16.98 \small(\textcolor{WildStrawberry}{+0.06}) & 63.67 \small(\textcolor{WildStrawberry}{-0.05})  \\

Non. Lin. $\Phi_{\boldsymbol{h}}^{t \rightarrow d}$ & 10.11 \small(\textcolor{WildStrawberry}{+1.27})  & 16.06 \small(\textcolor{WildStrawberry}{+1.69}) & 78.97 \small(\textcolor{WildStrawberry}{-3.70}) & 15.62 \small(\textcolor{WildStrawberry}{+1.56}) & 24.65 \small(\textcolor{WildStrawberry}{+3.17}) & 53.03 \small(\textcolor{WildStrawberry}{-1.85}) & 8.92 \small(\textcolor{Green}{-0.21}) & 16.52 \small(\textcolor{Green}{-0.30}) & 64.29 \small(\textcolor{Green}{+0.57})  \\
\bottomrule
\end{tabular}
\end{adjustbox}
\end{table*}

\begin{table*}[h]
\centering
\caption{Further ablation on proposed method in \textit{IMU} datasets for Activity and Step}
\begin{adjustbox}{width=\columnwidth,center}
\label{tab:performance_activity_ablation_2}
\renewcommand{\arraystretch}{1.1}
\begin{tabular}{@{}lllllllllll@{}}
\toprule
\multirow{2}{*}{Method} & \multicolumn{3}{l}{HHAR} & \multicolumn{3}{l}{USC} & \multicolumn{3}{l}{Clemson} \\ 
\cmidrule(r{15pt}){2-4}  \cmidrule(r{15pt}){5-7}  \cmidrule(r{15pt}){8-10} 
& Acc $\uparrow$ & W-F1 $\uparrow$ & F1 $\uparrow$ & Acc $\uparrow$ & W-F1 $\uparrow$ & F1 $\uparrow$ & MAPE $\downarrow$ & MAE $\downarrow$ & RMSE $\downarrow$ \\
\midrule
Ours & 70.67\small$\pm$0.06 & 67.74\small$\pm$0.29 & 68.79\small$\pm$0.25 & 52.21\small$\pm$1.09 & 48.64\small$\pm$1.52 & 48.22\small$\pm$1.11 & \textbf{5.16}\small$\pm$0.44 & \textbf{2.50}\small$\pm$0.23 & \textbf{4.65}\small$\pm$0.50 & \\
w/o $\Phi_{\boldsymbol{z}}^{t \rightarrow d}$ & 67.17 \small(\textcolor{WildStrawberry}{-3.50})  & 67.03 \small(\textcolor{WildStrawberry}{-0.71}) & 67.01 \small(\textcolor{WildStrawberry}{-1.78}) & 52.09 \small(\textcolor{WildStrawberry}{-0.12}) & 49.31 \small(\textcolor{Green}{+0.67}) & \textbf{49.25} \small(\textcolor{Green}{+1.03}) & 5.31 \small(\textcolor{WildStrawberry}{+0.15}) & 2.57 \small(\textcolor{WildStrawberry}{+0.07}) & 4.76 \small(\textcolor{WildStrawberry}{+0.11})  \\

\noalign{\vskip 1mm}
\hline  
\noalign{\vskip 1mm}

Lin. $\Phi_{\boldsymbol{h}}^{t \rightarrow d}$ & 68.13 \small(\textcolor{WildStrawberry}{-2.54}) & 68.23 \small(\textcolor{Green}{+0.49}) & 67.08 \small(\textcolor{WildStrawberry}{-1.71}) & 51.13 \small(\textcolor{WildStrawberry}{-1.08}) & 47.55 \small(\textcolor{WildStrawberry}{-1.09}) & 47.54 \small(\textcolor{WildStrawberry}{-0.68}) & 5.79 \small(\textcolor{WildStrawberry}{+0.63}) & 2.55 \small(\textcolor{WildStrawberry}{+0.05}) & 4.67 \small(\textcolor{WildStrawberry}{+0.02})  \\

Non. Lin. $\Phi_{\boldsymbol{h}}^{t \rightarrow d}$ & \textbf{71.13} \small(\textcolor{Green}{+0.46}) & \textbf{69.19} \small(\textcolor{Green}{+1.45}) & \textbf{70.25} \small(\textcolor{Green}{+1.46}) & \textbf{53.20} \small(\textcolor{Green}{+0.99}) & 48.27 \small(\textcolor{WildStrawberry}{-0.37}) & 49.01 \small(\textcolor{Green}{+0.79}) & 6.40 \small(\textcolor{WildStrawberry}{+1.24}) & 3.11 \small(\textcolor{WildStrawberry}{+0.61}) & 5.46 \small(\textcolor{WildStrawberry}{+0.81})  \\
\bottomrule
\end{tabular}
\end{adjustbox}
\end{table*}

\begin{table*}[h]
\centering
\caption{Further ablation on proposed method in \textit{ECG} and \textit{EEG} datasets for CVD and Sleep}
\begin{adjustbox}{width=\columnwidth,center}
\label{tab:performance_ecg_eeg_ablation_2}
\renewcommand{\arraystretch}{1.1}
\begin{tabular}{@{}lllllllllll@{}}
\toprule
\multirow{2}{*}{Method} & \multicolumn{3}{l}{Chapman} & \multicolumn{3}{l}{CPSC} & \multicolumn{3}{l}{Sleep} \\ 
\cmidrule(r{15pt}){2-4}  \cmidrule(r{15pt}){5-7}  \cmidrule(r{15pt}){8-10} 
& Acc $\uparrow$ & F1 $\uparrow$ & AUC $\uparrow$ & Acc $\uparrow$ & F1 $\uparrow$ & AUC $\uparrow$ & Acc $\uparrow$ & F1 $\uparrow$ & Kappa ($\kappa$) $\uparrow$ \\
\midrule
Ours & 87.21 \small$\pm$0.80 & 96.50 \small$\pm$0.21 & 85.30 \small$\pm$0.98 & 52.10\small$\pm$0.90 & 87.01 \small$\pm$1.10 & 51.26 \small$\pm$1.18 & 77.30 \small$\pm$1.04 & 68.05 \small$\pm$0.86 & 69.16 \small$\pm$1.32 \\ 
w/o $\Phi_{\boldsymbol{z}}^{t \rightarrow d}$ & 86.43 \small(\textcolor{WildStrawberry}{-0.78})  & 96.31 \small(\textcolor{WildStrawberry}{-0.19}) & 84.40 \small(\textcolor{WildStrawberry}{-0.90}) & 52.03 \small(\textcolor{WildStrawberry}{-0.07}) & 87.39 \small(\textcolor{Green}{+0.38}) & 51.10 \small(\textcolor{WildStrawberry}{-0.16}) & 77.98 \small(\textcolor{Green}{+0.68}) & 68.04 \small(\textcolor{WildStrawberry}{-0.01}) & 70.23 \small(\textcolor{Green}{+1.11})  \\

\noalign{\vskip 1mm}
\hline  
\noalign{\vskip 1mm}

Lin. $\Phi_{\boldsymbol{h}}^{t \rightarrow d}$ & 86.35 \small(\textcolor{WildStrawberry}{-0.86})  & 96.06 \small(\textcolor{WildStrawberry}{-0.44}) & 84.26 \small(\textcolor{WildStrawberry}{-1.04}) & 50.58 \small(\textcolor{WildStrawberry}{-1.52}) & 86.55 \small(\textcolor{WildStrawberry}{-0.46}) & 48.78 \small(\textcolor{WildStrawberry}{-1.48}) & 77.83 \small(\textcolor{Green}{+0.53}) & 68.61 \small(\textcolor{Green}{+0.56}) & 70.02 \small(\textcolor{Green}{+0.86})  \\
Non. Lin. $\Phi_{\boldsymbol{h}}^{t \rightarrow d}$ & \textbf{90.91} \small(\textcolor{Green}{+3.70})  & \textbf{97.96} \small(\textcolor{Green}{+1.46}) & \textbf{89.76} \small(\textcolor{Green}{+4.46}) & \textbf{56.30} \small(\textcolor{Green}{+4.20}) & \textbf{88.09} \small(\textcolor{Green}{+1.08}) & \textbf{53.44} \small(\textcolor{Green}{+2.18}) & \textbf{79.93} \small(\textcolor{Green}{+2.63}) & \textbf{70.27} \small(\textcolor{Green}{+2.22}) & \textbf{73.10} \small(\textcolor{Green}{+3.94})  \\
\bottomrule
\end{tabular}
\end{adjustbox}
\end{table*}

As can be seen in these tables, removing the embedding space mappers ($\Phi_{\boldsymbol{z}}^{t \rightarrow d}$) consistently leads to decreased performance across all datasets.
In a few cases where the performance appears slightly better, the improvement remains within the range of standard deviation and is not statistically significant.
These results support the role of embedding-level mapping in enhancing performance. 


\subsection{Latent space mappers}
When using latent space mappers, we employed lightweight nonlinear convolutional networks (see Section~\ref{appendix:Implementation_Details} for architectural details).
Here, we report results from replacing the mapper with either a simple linear layer (Lin. $\Phi_{\boldsymbol{z}}^{t \rightarrow d}$) or a nonlinear two-layer multilayer perceptron (MLP $\Phi_{\boldsymbol{z}}^{t \rightarrow d}$).
We present the results in Tables~\ref{tab:performance_ppg_ablation_2},~\ref{tab:performance_activity_ablation_2} and~\ref{tab:performance_ecg_eeg_ablation_2}.

As shown in the tables, replacing the convolutional mapper with a linear layer results in a noticeable performance drop, suggesting that the relationships between latent spaces are too complex for simple linear mappings.
Although a non-linear MLP improves performance on some datasets, its two-layer structure significantly increases the parameter count.
To ensure a fair comparison with prior work while avoiding added computational overhead, we use our lightweight convolutional architecture.

\subsection{Using all encoders instead of mappers}
We evaluate a variant of our method that directly uses the original representations from all encoders, rather than mapping the time-domain representations into other domains. This setup yields a 4--5\% performance improvement over our default approach.
However, it significantly increases the number of parameters at inference time, making it less efficient.
These results suggest a promising direction for future work: leveraging principled transformations can substantially boost performance, though careful trade-offs with inference cost should also be considered.

\subsection{Cross domain results}
Cross-domain evaluations are commonly used to assess the generalization of self-supervised learning methods across datasets~\cite{tf_consistent, dong2023simmtm}.
Following this practice, we pretrain each model on a dataset and fine tune it on another from a different domain, following the setup in~\cite{dong2023simmtm}.
Both supervised models in our experiments (FCN and ResNet) are initialized randomly and trained.
We present the result in Tables~\ref{tab:crossdomain_results1} and~\ref{tab:crossdomain_results2}.
Unlike linear probing, where only the linear classifier is trained with a limited data, fine-tuning updates the entire encoder. 
To ensure consistency, we use the same limited data size as in linear probing, avoiding settings that resemble supervised training with abundant labeled data.

\begin{table*}[h]
\caption{Performance comparison of methods for \textit{Activity} and \textit{Step} in cross domain settings}
\begin{adjustbox}{width=\columnwidth,center}
\label{tab:crossdomain_results1}
\renewcommand{\arraystretch}{1.3}
\begin{tabular}{@{}lllllllllll@{}}
\toprule
\multirow{2}{*}{Method} & \multicolumn{3}{l}{HHAR} & \multicolumn{3}{l}{USC} & \multicolumn{3}{l}{Clemson} \\
\cmidrule(r{15pt}){2-4}  \cmidrule(r{15pt}){5-7}  \cmidrule(r{15pt}){8-10}
& Acc $\uparrow$ & W-F1 $\uparrow$ & F1 $\uparrow$
& Acc $\uparrow$ & W-F1 $\uparrow$ & F1 $\uparrow$
& MAPE $\downarrow$ & MAE $\downarrow$ & RMSE $\downarrow$  \\
\midrule
\textit{Supervised} & & & &  \\
FCN & 74.21\small$\pm$1.56 & 72.88\small$\pm$2.06 & 71.58\small$\pm$1.81 & 48.87\small$\pm$0.74 & 46.02\small$\pm$0.95 & 45.33\small$\pm$0.82 & 5.02\small$\pm$0.26 & 2.86\small$\pm$0.15 & 4.05\small$\pm$0.13 \\
ResNet & 69.85\small$\pm$2.32 & 68.61\small$\pm$2.81 & 67.29\small$\pm$2.52 & 52.17\small$\pm$1.22 & 49.38\small$\pm$0.84 & 48.01\small$\pm$1.22 & 6.55\small$\pm$2.37 & 3.78\small$\pm$1.44 & 5.04\small$\pm$1.43 \\
\midrule
\textit{Self-Supervised} & & & & \\

SimCLR (in) & 40.55\small$\pm$0.62 & 39.21\small$\pm$0.64 & 39.41\small$\pm$0.66 & 29.16\small$\pm$0.69 & 29.02\small$\pm$0.67 & 28.99\small$\pm$0.79 & 8.70\small$\pm$0.22 & 4.36\small$\pm$0.13 & 6.30\small$\pm$0.24 \\
SimCLR (cross) & 41.25\small$\pm$0.62 & 39.75\small$\pm$0.64 & 40.05\small$\pm$0.66 & 29.75\small$\pm$0.69 & 29.55\small$\pm$0.67 & 29.45\small$\pm$0.79 & 8.55\small$\pm$0.22 & 4.28\small$\pm$0.13 & 6.18\small$\pm$0.24 \\

\noalign{\vskip 1mm}
\hline
\noalign{\vskip 1mm}

BYOL (in) & 49.64\small$\pm$2.48 & 48.63\small$\pm$2.75 & 48.02\small$\pm$2.59 & 28.40\small$\pm$1.23 & 28.23\small$\pm$1.42 & 28.23\small$\pm$0.96 & 9.35\small$\pm$0.19 & 4.72\small$\pm$0.12 & 6.79\small$\pm$0.24 \\
BYOL (cross) & 50.10\small$\pm$2.48 & 49.05\small$\pm$2.75 & 48.60\small$\pm$2.59 & 28.95\small$\pm$1.23 & 28.70\small$\pm$1.42 & 28.65\small$\pm$0.96 & 9.20\small$\pm$0.19 & 4.62\small$\pm$0.12 & 6.68\small$\pm$0.24 \\

\noalign{\vskip 1mm}
\hline
\noalign{\vskip 1mm}

VICReg (in) & 38.05\small$\pm$3.01 & 37.12\small$\pm$2.66 & 37.38\small$\pm$3.02 & 23.75\small$\pm$1.00 & 23.16\small$\pm$1.03 & 22.92\small$\pm$1.21 & 10.87\small$\pm$0.61 & 5.47\small$\pm$0.35 & 7.78\small$\pm$0.14 \\
VICReg (cross) & 38.55\small$\pm$3.01 & 37.55\small$\pm$2.66 & 37.85\small$\pm$3.02 & 24.30\small$\pm$1.00 & 23.70\small$\pm$1.03 & 23.40\small$\pm$1.21 & 10.70\small$\pm$0.61 & 5.35\small$\pm$0.35 & 7.65\small$\pm$0.14 \\

\noalign{\vskip 1mm}
\hline
\noalign{\vskip 1mm}

Barlow Twins (in) & 38.97\small$\pm$0.65 & 37.75\small$\pm$1.00 & 38.21\small$\pm$1.12 & 27.24\small$\pm$0.19 & 26.84\small$\pm$0.20 & 26.25\small$\pm$0.77 & 9.89\small$\pm$0.35 & 4.95\small$\pm$0.15 & 7.03\small$\pm$0.21 \\
Barlow Twins (cross) & 39.60\small$\pm$0.65 & 38.25\small$\pm$1.00 & 38.85\small$\pm$1.12 & 27.85\small$\pm$0.19 & 27.30\small$\pm$0.20 & 26.80\small$\pm$0.77 & 9.72\small$\pm$0.35 & 4.85\small$\pm$0.15 & 6.92\small$\pm$0.21 \\

\noalign{\vskip 1mm}
\hline
\noalign{\vskip 1mm}

CLIP (in) & 43.78\small$\pm$0.89 & 42.53\small$\pm$0.90 & 43.07\small$\pm$0.98 & 25.55\small$\pm$0.63 & 25.78\small$\pm$1.25 & 25.17\small$\pm$0.75 & 8.52\small$\pm$0.46 & 4.26\small$\pm$0.23 & 6.73\small$\pm$0.63 \\
CLIP (cross) & 44.25\small$\pm$0.89 & 42.95\small$\pm$0.90 & 43.55\small$\pm$0.98 & 26.10\small$\pm$0.63 & 26.25\small$\pm$1.25 & 25.65\small$\pm$0.75 & 8.35\small$\pm$0.46 & 4.15\small$\pm$0.23 & 6.60\small$\pm$0.63 \\

\noalign{\vskip 1mm}
\hline
\noalign{\vskip 1mm}

TS-TCC (in) & \underline{68.56}\small$\pm$1.19 & \underline{66.90}\small$\pm$1.22 & \underline{68.10}\small$\pm$1.30 & 33.61\small$\pm$0.72 & \underline{33.11}\small$\pm$1.09 & 33.91\small$\pm$0.79 & \underline{5.61}\small$\pm$0.15 & \underline{2.70}\small$\pm$0.06 & \underline{4.69}\small$\pm$0.38 \\
TS-TCC (cross) & 69.10\small$\pm$1.19 & 67.35\small$\pm$1.22 & 68.60\small$\pm$1.30 & 34.15\small$\pm$0.72 & 33.60\small$\pm$1.09 & 34.30\small$\pm$0.79 & 5.50\small$\pm$0.15 & 2.62\small$\pm$0.06 & 4.60\small$\pm$0.38 \\

\noalign{\vskip 1mm}
\hline
\noalign{\vskip 1mm}

SimMTM (in) & 44.78\small$\pm$0.62 & 42.48\small$\pm$0.37 & 43.60\small$\pm$0.62 & 22.34\small$\pm$0.28 & 25.68\small$\pm$0.41 & 29.72\small$\pm$1.78 & 8.77\small$\pm$0.18 & 4.61\small$\pm$0.32 & 6.90\small$\pm$0.18 \\
SimMTM (cross) & 45.52\small$\pm$0.54 & 43.01\small$\pm$0.41 & 44.10\small$\pm$0.68 & 22.90\small$\pm$0.28 & 26.10\small$\pm$0.41 & 30.05\small$\pm$1.78 & 8.60\small$\pm$0.18 & 4.50\small$\pm$0.32 & 6.78\small$\pm$0.18 \\

\noalign{\vskip 1mm}
\hline
\noalign{\vskip 1mm}

TF-C (in) & 31.13\small$\pm$0.42 & 30.57\small$\pm$0.40 & 31.00\small$\pm$0.31 & 30.78\small$\pm$0.39 & 28.16\small$\pm$0.23 & 30.82\small$\pm$1.41 & 12.47\small$\pm$0.72 & 6.31\small$\pm$0.37 & 7.93\small$\pm$0.30 \\
TF-C (cross) & 31.70\small$\pm$0.42 & 31.05\small$\pm$0.40 & 31.55\small$\pm$0.31 & 31.30\small$\pm$0.39 & 28.75\small$\pm$0.23 & 31.25\small$\pm$1.41 & 12.25\small$\pm$0.72 & 6.15\small$\pm$0.37 & 7.75\small$\pm$0.30 \\

\noalign{\vskip 1mm}
\hline
\noalign{\vskip 1mm}

TS2Vec (in) & 67.13\small$\pm$0.11 & 65.56\small$\pm$0.21 & 64.13\small$\pm$0.21 & \underline{35.40}\small$\pm$0.96 & 32.17\small$\pm$1.26 & \underline{35.47}\small$\pm$1.42 & 5.92\small$\pm$0.93 & 3.01\small$\pm$0.28 & 5.02\small$\pm$0.42 \\
TS2Vec (cross) & 67.70\small$\pm$0.11 & 66.05\small$\pm$0.21 & 64.70\small$\pm$0.21 & 36.00\small$\pm$0.96 & 32.75\small$\pm$1.26 & 36.05\small$\pm$1.42 & 5.85\small$\pm$0.93 & 2.95\small$\pm$0.28 & 4.95\small$\pm$0.42 \\

\noalign{\vskip 1mm}
\hline
\noalign{\vskip 1mm}

Ours (in) & 70.67\small$\pm$0.06 & 67.74\small$\pm$0.29 & 68.79\small$\pm$0.25 & 52.21\small$\pm$1.09 & 48.64\small$\pm$1.52 & 48.22\small$\pm$1.11 & 5.16\small$\pm$0.44 & 2.50\small$\pm$0.13 & 4.65\small$\pm$0.45 \\
Ours (cross) & \textbf{71.20}\small$\pm$0.06 & \textbf{68.20}\small$\pm$0.29 & \textbf{69.30}\small$\pm$0.25 & \textbf{53.10}\small$\pm$1.09 & \textbf{49.20}\small$\pm$1.52 & \textbf{48.80}\small$\pm$1.11 & \textbf{5.05}\small$\pm$0.44 & \textbf{2.40}\small$\pm$0.13 & \textbf{4.55}\small$\pm$0.45 \\

\bottomrule
\end{tabular}
\end{adjustbox}
\end{table*}

For experiments regarding the activity recognition and step counting tasks (Table~\ref{tab:crossdomain_results1}), we have first used electrocardiogram signals to pretrain models and then fine tune on the specific dataset.
Each method includes two rows: the first shows in-domain results (also reported in the main manuscript), and the second shows cross-domain performance.
Results are organized top-to-bottom per method to facilitate easier comparison.

For the cardiovascular disease classification (CVD) task (Table~\ref{tab:crossdomain_results2}), we pretrain models on EEG signals and fine-tune them on ECG datasets.
For sleep stage classification, we pretrain on both ECG datasets and fine-tune on a small subset of the EEG dataset.
The results are summarized in Table~\ref{tab:crossdomain_results2}.

\clearpage

\begin{table*}[h]
\caption{Performance comparison of methods for \textit{CVD} and \textit{Sleep} in cross domain settings}
\begin{adjustbox}{width=\columnwidth,center}
\label{tab:crossdomain_results2}
\renewcommand{\arraystretch}{1.3}
\begin{tabular}{@{}lllllllllll@{}}
\toprule
\multirow{2}{*}{Method} & \multicolumn{3}{l}{Chapman} & \multicolumn{3}{l}{CPSC} & \multicolumn{3}{l}{Sleep} \\ 
\cmidrule(r{15pt}){2-4}  \cmidrule(r{15pt}){5-7}  \cmidrule(r{15pt}){8-10}  
& Acc $\uparrow$ & AUC $\uparrow$ & F1 $\uparrow$ & Acc $\uparrow$ & AUC $\uparrow$ & F1 $\uparrow$ & Acc $\uparrow$ & W-F1 $\uparrow$ & Kappa $\uparrow$  \\
\midrule
\textit{Supervised} & & & &  \\
FCN & 84.63\small$\pm$2.13 & 95.40\small$\pm$0.57 & 82.41\small$\pm$2.40 & 63.64\small$\pm$1.12 & 91.30\small$\pm$0.02 & 60.43\small$\pm$1.04 & 71.98\small$\pm$0.86 & 63.33\small$\pm$0.84 & 62.01\small$\pm$1.30  \\
ResNet & 93.16\small$\pm$0.41 & 98.59\small$\pm$0.05 & 92.02\small$\pm$0.42 & 75.21\small$\pm$1.73 & 95.02\small$\pm$0.03 & 71.70\small$\pm$1.90 & 76.94\small$\pm$0.97 & 67.52\small$\pm$1.95 & 69.14\small$\pm$0.61 \\
\midrule

SimCLR (in)  & 75.28\small$\pm$0.57 & 93.55\small$\pm$0.25 & 74.04\small$\pm$0.50 & 50.10\small$\pm$0.41 & 87.20\small$\pm$0.07 & 50.10\small$\pm$0.24 & 72.45\small$\pm$2.32 & 58.93\small$\pm$1.59 & 59.47\small$\pm$3.20 &\\

SimCLR (cross) & 76.58\small$\pm$0.55 & 93.92\small$\pm$0.22 & 75.39\small$\pm$0.48 & 51.03\small$\pm$0.40 & \underline{87.48}\small$\pm$0.06 & 50.25\small$\pm$0.22 & 73.56\small$\pm$2.25 & 59.84\small$\pm$1.55 & 60.41\small$\pm$3.10\\

\noalign{\vskip 1mm}
\hline  
\noalign{\vskip 1mm}

BYOL (in)  & 77.08\small$\pm$0.40 & 93.74\small$\pm$0.18 & 75.80\small$\pm$0.35 & 52.90\small$\pm$0.30 & 87.05\small$\pm$0.22 & 50.89\small$\pm$0.38 & 70.77\small$\pm$0.27 & 58.23\small$\pm$0.55 & 55.90\small$\pm$1.20 \\

BYOL (cross) & 77.10\small$\pm$0.38 & 94.81\small$\pm$0.16 & 77.18\small$\pm$0.42 & \underline{52.95}\small$\pm$0.28 & {87.32}\small$\pm$0.20 & \underline{51.35}\small$\pm$0.36 & 71.82\small$\pm$0.25 & 59.29\small$\pm$0.52 & 56.80\small$\pm$1.15 \\

\noalign{\vskip 1mm}
\hline  
\noalign{\vskip 1mm}

VICReg (in)  & 70.10\small$\pm$1.90 & 89.35\small$\pm$0.93 & 67.84\small$\pm$1.79 & 46.21\small$\pm$1.29 & 84.70\small$\pm$0.50 & 42.51\small$\pm$0.96 & 68.72\small$\pm$1.03 & 57.24\small$\pm$1.04 & 57.13\small$\pm$1.42
\\

VICReg (cross) & 71.31\small$\pm$1.83 & 89.78\small$\pm$0.90 & 68.75\small$\pm$1.70 & 47.38\small$\pm$1.26 & 85.09\small$\pm$0.47 & 43.25\small$\pm$0.91 & 69.75\small$\pm$1.00 & 58.10\small$\pm$1.01 & 58.06\small$\pm$1.38 \\

\noalign{\vskip 1mm}
\hline  
\noalign{\vskip 1mm}

Barlow Twins (in)  & 72.43\small$\pm$1.45 & 91.17\small$\pm$0.60 & 70.42\small$\pm$1.53 & 48.67\small$\pm$0.51 & 85.78\small$\pm$0.19 & 44.57\small$\pm$0.53 & 70.10\small$\pm$0.62 & 57.72\small$\pm$0.81 & 57.88\small$\pm$0.82 \\

Barlow Twins (cross) & 73.21\small$\pm$1.41 & 91.61\small$\pm$0.57 & 71.38\small$\pm$1.50 & 49.32\small$\pm$0.48 & 86.14\small$\pm$0.18 & 45.33\small$\pm$0.50 & 71.13\small$\pm$0.60 & 58.52\small$\pm$0.78 & 58.71\small$\pm$0.79 \\

\noalign{\vskip 1mm}
\hline  
\noalign{\vskip 1mm}

CLIP (in)  & 82.98\small$\pm$0.96 & 95.15\small$\pm$0.42 & 81.00\small$\pm$1.03 & 50.01\small$\pm$0.89 & 86.40\small$\pm$0.32 & 47.99\small$\pm$0.89 & 73.16\small$\pm$0.81 & 62.06\small$\pm$0.91 & 63.75\small$\pm$1.23 \\

CLIP (cross) & 83.69\small$\pm$0.94 & 95.39\small$\pm$0.39 & 81.85\small$\pm$1.01 & 51.13\small$\pm$0.85 & 86.61\small$\pm$0.31 & 48.88\small$\pm$0.86 & 74.25\small$\pm$0.78 & 63.10\small$\pm$0.89 & 64.80\small$\pm$1.20 \\

\noalign{\vskip 1mm}
\hline  
\noalign{\vskip 1mm}

TS-TCC (in)  & 73.50\small$\pm$0.55 & 90.65\small$\pm$0.07 & 71.10\small$\pm$0.57 & 51.59\small$\pm$1.22 & 86.32\small$\pm$0.16 & 50.27\small$\pm$1.32 & 62.80\small$\pm$1.13 & 52.43\small$\pm$1.05 & 48.98\small$\pm$1.68 \\

TS-TCC (cross) & 74.66\small$\pm$0.53 & 91.00\small$\pm$0.06 & 72.10\small$\pm$0.55 & 52.35\small$\pm$1.18 & 86.51\small$\pm$0.15 & 51.17\small$\pm$1.29 & 63.78\small$\pm$1.10 & 53.42\small$\pm$1.03 & 49.97\small$\pm$1.65 \\

\noalign{\vskip 1mm}
\hline  
\noalign{\vskip 1mm}

SimMTM (in)  & 84.29\small$\pm$1.29 & 95.87\small$\pm$0.18 & 83.31\small$\pm$1.25 & 51.70\small$\pm$0.23 & 87.08\small$\pm$0.21 & 50.62\small$\pm$0.55 & {74.69}\small$\pm$1.84 & {63.53}\small$\pm$1.21 & {65.31}\small$\pm$2.76 \\

SimMTM (cross) & 85.21\small$\pm$1.25 & 96.03\small$\pm$0.17 & 84.20\small$\pm$1.20 & 52.34\small$\pm$0.22 & 87.33\small$\pm$0.20 & 51.52\small$\pm$0.53 & {75.88}\small$\pm$1.78 & {64.60}\small$\pm$1.18 & {66.31}\small$\pm$2.71 \\

\noalign{\vskip 1mm}
\hline  
\noalign{\vskip 1mm}

TF-C (in)  & {85.84}\small$\pm$0.39 & {96.10}\small$\pm$0.10 & 84.71\small$\pm$0.40 & 47.86\small$\pm$0.69 & 86.27\small$\pm$0.05 & 45.42\small$\pm$0.66 & 64.50\small$\pm$1.80 & 56.77\small$\pm$2.21 & 52.61\small$\pm$2.41 \\

TF-C (cross) & {86.67}\small$\pm$0.37 & {96.28}\small$\pm$0.09 & 85.43\small$\pm$0.38 & 49.31\small$\pm$0.67 & 86.55\small$\pm$0.04 & 46.21\small$\pm$0.64 & 65.65\small$\pm$1.75 & 57.86\small$\pm$2.17 & 53.72\small$\pm$2.35 \\

\noalign{\vskip 1mm}
\hline  
\noalign{\vskip 1mm}

TS2Vec (in)  & 78.87\small$\pm$1.03 & 90.23\small$\pm$0.24 & 81.32\small$\pm$0.47 & 48.73\small$\pm$0.85 & 85.49\small$\pm$0.37 & 46.57\small$\pm$1.10 & 65.71\small$\pm$1.06 & 55.32\small$\pm$1.77 & 56.81\small$\pm$1.90 \\

TS2Vec (cross) & 80.03\small$\pm$1.00 & 90.48\small$\pm$0.23 & 82.13\small$\pm$0.45 & 49.61\small$\pm$0.82 & 85.76\small$\pm$0.35 & 47.28\small$\pm$1.07 & 66.94\small$\pm$1.04 & 56.34\small$\pm$1.74 & 57.84\small$\pm$1.87 \\

\noalign{\vskip 1mm}
\hline  
\noalign{\vskip 1mm}

Ours (in)  & \underline{87.21}\small$\pm$0.80 & \underline{96.50}\small$\pm$0.21 & \underline{85.30}\small$\pm$0.98 & 52.10\small$\pm$0.90 & 87.11\small$\pm$0.40 & 51.26\small$\pm$1.18 & \underline{77.30}\small$\pm$1.04 & \underline{68.05}\small$\pm$0.86 & \underline{69.16}\small$\pm$1.32 & \\

Ours (cross) & \textbf{88.03}\small$\pm$0.78 & \textbf{96.78}\small$\pm$0.20 & \textbf{86.01}\small$\pm$0.95 & \textbf{53.11}\small$\pm$0.87 & \textbf{88.03}\small$\pm$0.48 & \textbf{52.02}\small$\pm$1.14 & \textbf{78.21}\small$\pm$1.01 & \textbf{69.01}\small$\pm$0.83 & \textbf{70.15}\small$\pm$1.28 \\

\bottomrule
\end{tabular}
\end{adjustbox}
\end{table*}

As shown in Tables~\ref{tab:crossdomain_results1} and~\ref{tab:crossdomain_results2}, our method demonstrates strong generalization across domains, achieving the best performance in 17 out of 18 metrics across 6 datasets.
This is particularly important because our approach employs specific transformations, such as Fourier and wavelet projections, which are inherently sensitive to the signal’s sampling rate.

Despite variations in sampling rates between source and target datasets, our method consistently outperforms others.
While similar resilience to sampling rate differences has been observed in prior works that employs frequency-domain representations~\cite{tf_consistent}, we hypothesize that the adaptability of our method stems from fine-tuning both the main encoder and the latent space mappers, which enables adaptation to cross-domain evaluation similar to previous methods.

\subsection{Comparison with larger models}
\label{appendix:foundation_models}
In our experiments, we aimed to ensure fair comparisons across methods by controlling for model capacity and training data.  
Therefore, we used the same or closely matched backbone architectures with the same amount of training data across methods, allowing us to isolate and evaluate the effect of the representation learning strategy itself rather than differences in model or data scale.  

Foundation models, while powerful, involve substantially larger architectures and extensive pretraining on massive datasets. 
As such, a direct comparison in our experiments is not straightforward.
Nonetheless, for completeness, we conducted experiments with CBraMod~\cite{wang2025cbramod}, one of the most recent of these foundation models for brain signals.
Specifically, we used the publicly available pretrained weights and evaluated on the Sleep-EDF dataset.
Features were extracted using the pretrained model, followed by either (i) training only a linear classifier or (ii) fine-tuning the full network on the same training split.  
We have reported the results in Table~\ref{tab:cbramod_comparison}.
\begin{table}[h]
    \centering
    \caption{Comparison with the pretrained CBraMod on Sleep-EDF.}
    \renewcommand{\arraystretch}{1.1}
    \label{tab:cbramod_comparison}
    \begin{tabular}{lcc}
        \hline
        \textbf{Method} & \textbf{Linear Evaluation} & \textbf{Fine-tuning} \\ \hline
        CBraMod (pretrained) & 65.03 $\pm$ 0.63 & \textbf{72.98 $\pm$ 0.47} \\
        Ours & \textbf{69.19 $\pm$ 1.32} & 71.13 $\pm$ 0.84 \\ \hline
    \end{tabular}
\end{table}
CBraMod employs a transformer layer for sequence-to-sequence sleep staging, whereas we applied a linear classifier for consistency with other baselines.  
It is also worth noting that CBraMod was not originally trained on single-channel EEG, which may partly explain the lower evaluation performance.  
While fine-tuning increases the performance of CBraMod more compared to ours, it is important to note that it has a larger parameter count than our backbone, giving it greater capacity to adapt when all weights are updated according to the task.
This makes direct comparison under fine-tuning less informative, as the performance difference can largely reflect model size rather than the quality of the learned representations.

By contrast, linear evaluation offers a fair basis for comparison, as only a lightweight classifier is trained.  
Our results show that, despite its smaller design, our approach remains competitive.  

\subsection{Comparison with heavy specialized augmentations}
\label{appendix:heavy_aug}
To further assess the performance of our method, we compared it against augmentation-heavy baselines, including TimesURL~\cite{TimesURL} and Finding Order in Chaos~\cite{demirel2023chaos}, which introduces frequency domain based mixup.  
Table~\ref{tab:timesurl_comparison} reports results.  
For Finding Order in Chaos (FOC), we reproduced the results on SPC12 (where dataset splits differ slightly from our setup), and report the original numbers from the paper for SPC22 and DaLiA.  
Our method consistently outperforms augmentation-heavy baselines, with improvements of approximately 8--9\% across all datasets.  
\begin{table*}[h]
\caption{Performance comparison of our method with augmentation-heavy baselines}
\begin{adjustbox}{width=\columnwidth,center}
\label{tab:timesurl_comparison}
\renewcommand{\arraystretch}{1.1}
\begin{tabular}{@{}lllllllllll@{}}
\toprule
\multirow{2}{*}{Method} & \multicolumn{3}{l}{IEEE SPC12} & \multicolumn{3}{l}{IEEE SPC22} & \multicolumn{3}{l}{DaLiA} \\ 
\cmidrule(r{15pt}){2-4}  \cmidrule(r{15pt}){5-7}  \cmidrule(r{15pt}){8-10}  
& MAE $\downarrow$ & RMSE $\downarrow$ & $\rho$ $\uparrow$ & MAE $\downarrow$ & RMSE $\downarrow$ & $\rho$  $\uparrow$ & MAE $\downarrow$ & RMSE $\downarrow$ & $\rho$ $\uparrow$ \\
\midrule
\textit{Supervised} & & & &  \\
FCN & 15.13\small$\pm$0.50 & 21.63\small$\pm$0.48 & 52.09\small$\pm$5.43 & 16.57\small$\pm$0.91 & 26.20\small$\pm$0.60 & 55.98\small$\pm$0.78 & 12.45\small$\pm$0.12 & 18.35\small$\pm$0.24 & 56.98\small$\pm$0.78  \\
ResNet & 7.08\small$\pm$0.20 & 13.60\small$\pm$0.38 & 79.60\small$\pm$1.10 & 9.90\small$\pm$1.47 & 16.67\small$\pm$1.60 & 67.58\small$\pm$2.98 & 5.50\small$\pm$0.05 & 10.84\small$\pm$0.03 & 82.10\small$\pm$0.06 \\
\midrule
\textit{Self-Supervised} & & & & \\
SimCLR & 12.42\small$\pm$0.05 & 20.96\small$\pm$0.30 & 73.62\small$\pm$0.52 & 16.41\small$\pm$0.22 & 22.62\small$\pm$0.39 & 52.16\small$\pm$1.12 & 16.88\small$\pm$0.19 & 22.64\small$\pm$0.22 & 56.37\small$\pm$0.21 \\
TimesURL & 13.40\small$\pm$0.42 & 19.85\small$\pm$0.61 & 75.13\small$\pm$1.22 & 17.92\small$\pm$0.88 & 24.73\small$\pm$1.25 & 50.10\small$\pm$1.97 & 14.62\small$\pm$0.31 & 21.57\small$\pm$0.44 & 60.17\small$\pm$0.11 \\
FOC & 11.15\small$\pm$0.34 & 17.89\small$\pm$0.83 & 77.43\small$\pm$0.30 & \textbf{12.25\small$\pm$0.47} & \textbf{18.20\small$\pm$0.61} & \textbf{57.13\small$\pm$0.42} & 10.57\small$\pm$0.55 & 20.37\small$\pm$0.73 & 62.83\small$\pm$0.22 \\
Ours & \textbf{8.84}\small$\pm$0.50 & \textbf{14.37}\small$\pm$0.95 & \textbf{82.67}\small$\pm$1.30 & 14.06\small$\pm$1.09 & 21.48\small$\pm$2.01 & 54.88\small$\pm$1.89 & \textbf{9.13}\small$\pm$0.20 & \textbf{16.92}\small$\pm$0.56 & \textbf{63.72}\small$\pm$0.06 \\
\bottomrule
\end{tabular}
\end{adjustbox}
\end{table*}
These results highlight that our approach achieves stronger performance compared to methods based on heavy augmentations.  
While augmentations can be beneficial for certain datasets, their generalization is often limited due to the diverse characteristics of signals and the varying invariances required across tasks.  

\subsection{Significance analysis}
In our experiments, some methods achieve strong results on specific datasets; however, our approach consistently ranks highest across tasks.
For instance, TF-C performs well on ECG (ranking second to ours in Table~\ref{tab:performance_ecg_eeg}), but its accuracy falls to about 30\% on the activity recognition (Table~\ref{tab:performance_imu}).

\begin{figure}[b]
    \centering
    \includegraphics[width=\linewidth]{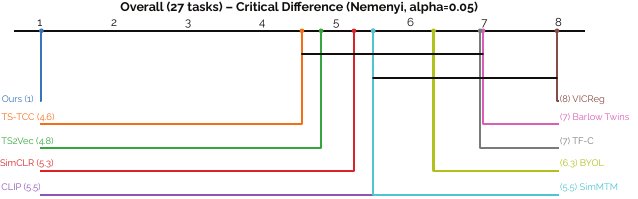}
    \caption{\label{fig:CD}The critical diagram for all tasks.
    Numbers show average ranks (lower is better); horizontal bars connect methods not significantly different. Our method achieves the best average rank.
    }
    \label{fig:placeholder}
\end{figure}

We evaluate significance across 27 tasks (3 datasets × 3 metrics for HR, 3×3 for Activity/Step, 3×3 for CVD/Sleep).
After ranking, we run the Friedman test and apply Nemenyi post-hoc comparisons. 
The critical difference diagram is given in Figure~\ref{fig:CD}.
\vspace{-5mm}

\clearpage

\section{Experiments}
\label{appendix:experiments}
Here, we give a detailed description of datasets, architectures, metrics, and training details for our experiments. 
We performed our experiments on NVIDIA GeForce RTX 4090 GPUs, involving training with three random seeds for all datasets, totaling approximately 680 GPU hours including ablation.
All experiments fit within 24 GB of GPU memory, without requiring excessive computational resources.
We reported the mean of three runs with the standard deviation.

\subsection{Datasets}
\label{appendix:datasets}
In this section, we give details about the datasets that are used during our experiments.

\subsubsection{Human Activity Recognition and Step counting}

\paragraph{Clemson}
The Clemson dataset has 30 participants (15 males, 15 females), where each participant wore three Shimmer3 sensors.
We used the IMU sensor readings from non-dominant wrists to predict step count where each sensor recorded accelerometer and gyroscope data at 15 Hz. 
We calculated the total magnitude of the accelerometer and fed it to the model as a pre-processing without any filtering.
We used window lengths of 32 seconds without an overlap in the regular walking setting.
We conducted 10-fold cross-validation, with each fold consisting of 3 subjects.
Pre-training is performed using 9 folds, with the remaining fold held out for testing. The test fold is not used during either pre-training or linear fine-tuning.

\paragraph{HHAR}
Heterogeneity Dataset for Human Activity Recognition (HHAR) is collected by nine subjects within an age range of 25 to 30 performing six daily living activities with eight different smartphones---Although HHAR includes data from smartwatches as well, we use data from smartphones---that were kept in a tight pouch and carried by the users around their waists~\cite{hhar}.
Subjects then perform 6 activities: ‘bike’, ‘sit’, ‘stairs down’, ‘stairs up’, ‘stand’, and ‘walk’.
Due to variant sampling frequencies of smart devices used in HHAR dataset, we downsample the readings to 50\,Hz and apply 100 (two seconds) and 50 as sliding window length with step size, the windows are normalized to zero mean with unit standard deviation.
We used the first four subjects (i.e., a, b, c, d) as source domains.

\paragraph{USC}
USC human activity dataset (USC-HAD) is composed of 14 subjects (7 male, 7 female, aged 21
to 49 with a mean of 30.1) executing 12 activities with a sensor on the front right hip.
The data dimension is six (3-axis accelerometer, 3-axis gyroscope) and the sample rate is 100\,Hz.
12 activities include walking forward, walking left, walking right, walking upstairs, walking
downstairs, running forward, jumping up, sitting, standing, sleeping, elevator up, and elevator down.
We used the pre-processing technique with a smaller window size such that the input contains six channels with 100 features (it is sampled in a sliding window of 1 second and 50\% overlap,
resulting in 100 features for each window).
The same normalization is also applied to windows before feeding to models.
We used the same setup with UCIHAR while source subjects are chosen as the last four this time.

\subsubsection{Heart Rate Prediction}

\paragraph{IEEE SPC}  This competition provided a training dataset of 12 subjects (SPC12) and a test dataset of 10 subjects~\cite{DeepPPG}. 
The IEEE SPC dataset overall has 22 recordings of 22 subjects, ages ranging from 18 to 58 performing three different activities~\cite{Binary_CorNet}. 
Each recording has sampled data from three accelerometer signals and two PPG signals along with the sampled ECG data and the sampling frequency is 125\,Hz. 
All these recordings were recorded from the wearable device placed on the wrist of each individual. 
All recordings were captured with a 2-channel pulse oximeter with green LEDs, a tri-axial accelerometer, and a chest ECG for the ground-truth HR estimation. 
During our experiments, we used PPG channels. 
We choose the first five subjects of SPC12 as source domains similar to \textit{activity recognition} setup while the last six subjects of SPC22 are used for source domains to prevent overlapping subjects with SPC12.

\paragraph{DaLiA} 
PPG dataset for motion compensation and heart rate estimation in Daily Life Activities (DaLiA) was recorded from 15 subjects (8 females, 7 males, mean age of $30.6$), where each recording was approximately two hours long. 
PPG signals were recorded while subjects went through different daily life activities, for instance sitting, walking, driving, cycling, working, and so on. 
PPG signals were recorded at a sampling rate of 64\,Hz. 
The first five subjects are used as source domains.

All PPG datasets are standardized as follows.
Initially, a fourth-order Butterworth bandpass filter with a frequency range of 0.5--4\,Hz is applied to PPG signals. 
Subsequently, a sliding window of 8 seconds with 2-second shifts is employed for segmentation, followed by z-score normalization of each segment. 
Lastly, the signal is resampled to a frequency of 25\,Hz for each segment.

\subsubsection{Cardiovascular disease (CVD) classification}

\paragraph{CPSC}
China Physiological Signal Challenge 2018 (CPSC2018), held during the 7th International Conference on Biomedical Engineering and Biotechnology in Nanjing, China. 
This dataset consists of 6,877 (male: 3,699; female: 3,178) and 12 lead ECG recordings lasting from 6 seconds to 60 seconds with 500\,Hz.
We use the original labelling~\cite{CPSC} with one normal and eight abnormal types as follows: atrial fibrillation, first-degree atrioventricular block, left bundle branch block, right bundle branch block, premature atrial contraction, premature ventricular contraction, ST-segment depression, ST-segment elevated.
We resampled recordings to 100\,Hz and excluded recordings of less than 10 seconds.

\paragraph{Chapman}
Chapman University, Shaoxing People’s Hospital (Chapman) ECG dataset which provides 12-lead ECG with 10 seconds of a sampling rate of 500\,Hz. 
The recordings are downsampled to 100\,Hz, resulting in each ECG frame consisting of 1000 samples.
The labeling setup follows the same approach as in~\cite{chapman} with four classes: atrial fibrillation, GSVT, sudden bradycardia, and sinus rhythm. 
The ECG frames are normalized to have a mean of 0 and scaled to have a standard deviation of 1.
We split the dataset to 80--20\% for training and testing as suggested in~\cite{chapman}.

We choose leads I, II, III, and V2 during our experiments for both ECG datasets.
We followed a similar setup with prior works~\cite{CLOCS} and considered each dataset as a single domain different from previous tasks.
The fine-tuning of the linear layer, which is added to the frozen pre-trained encoder, is performed with 80\% of the same domain.

\subsubsection{Sleep stage classification}
We used the Sleep-EDF dataset which has five classes: wake (W), three different non-rapid eye movements (N1, N2, N3), and rapid eye movement (REM).
The dataset includes whole-night PSG sleep recordings, where we used a single EEG channel (i.e., Fpz-Cz) with a sampling rate of 100\,Hz.
We followed the same data split as TSTCC~\citep{tstcc}, with no additional pre-processing. 
The only difference is that, for linear probing, we used a random 10\% subset of the unseen data rather than the full set, reflecting our setup where labeled data is significantly smaller than unlabeled data.

\subsection{Baselines}
\label{appendix:baselines}

\subsubsection{Supervised}
\paragraph{FCN}
We use a 1D Fully Convolutional Network (FCN) that processes multichannel temporal inputs.
The model consists of three convolutional layers with increasing filter sizes (32, 64, and 128), each followed by max pooling operations to progressively reduce the temporal resolution. 
A final linear layer maps the output to class logits. 
We chose this architecture for the supervised and self-supervised learning paradigms as it was widely used before in the literature~\cite{demirel2023chaos}.

\paragraph{ResNet}
We use the same backbone as in the self-supervised methods for the supervised baseline, integrating a linear layer and training the model from scratch using random initialization.

We perform a grid search over key hyperparameters for both networks, focusing on learning rate and batch size.
The learning rate is initialized at 1e-3 and reduced by half if validation performance does not improve for 15 epochs.
The batch size is fixed at 64.

\subsubsection{Self-Supervised}
\label{appendix:baselines_SSL}

\subsubsection*{Fundamentals}
\paragraph{SimCLR}
SimCLR~\cite{simclr} introduces a contrastive learning framework for self-supervised visual representation learning. 
The method relies on maximizing agreement between differently augmented views of the same image via a contrastive loss in the latent space. 
We follow the previous implementations of SimCLR for time series~\cite{demirel2023chaos, KDD_paper}.

\paragraph{BYOL}
For the BYOL implementation, the exponential moving average parameter is set to 0.996 where the projector size is set to 128.
We set the learning rate to 0.03 similar to other SSL techniques.
Following the original implementation, we use a weight decay parameter of $1.5e-6$.

\paragraph{VICReg}
We follow the original implementation and set the coefficients for each loss term to 25 ($\lambda$), 25 ($\mu$), and 1 ($\nu$), corresponding to the invariance, variance, and covariance terms, respectively.
Although we conducted a search for these loss term values, no performance enhancements were detected across the tasks.

\begin{equation}
    \ell = \lambda \left[ s(z,z^{\prime})\right] + \mu \left[ v(z) + v(z^\prime) \right] + \nu \left[  c(z) + c(z^{\prime}) \right],
\end{equation}

where $s$ is the mean-squared Euclidean distance, $v$ is a hinge function on the standard deviation of the embeddings along the batch dimension, $c$ is the covariance regularization term as the sum of the squared off-diagonal coefficients

\paragraph{Barlow Twins}
Barlow Twins~\cite{barlow_twins, barlow_real} presents an objective function that naturally avoids collapse for SSL by measuring the cross-correlation matrix between the outputs of two identical networks fed with augmented versions of a sample, and making it as close to the identity matrix as possible. 
This causes the embedding vectors of augmented versions of a sample to be similar, while minimizing the redundancy between the components of these vectors.
Following the original implementation, we applied batch normalization to the extracted embeddings and set the hyperparameter $\lambda$ coefficient (in Equation~\ref{eq:barlow_twins}) to 0.005.

\begin{equation}\label{eq:barlow_twins}
\mathcal{L} = \sum_{i} (1 - C_{ii})^2 + \lambda \sum_{i} \sum_{j \neq i} C_{ij}^2,
\end{equation}

where \( C \) is the cross-correlation matrix computed between the two sets of normalized embeddings.

\paragraph{CLIP}
Contrastive Language–Image Pretraining (CLIP)~\cite{CLIP} learns joint representations by aligning paired image and text embeddings through contrastive learning.
To compare our method with this, we adapted a similar strategy for time series by treating the time-domain signal and its Fourier-transformed version as two modalities.
Specifically, we applied a CLIP-style objective by training separate encoders for the time and frequency domains and maximizing the similarity between their paired embeddings.
During inference, we only used time encoder with linear probing.
This setup mirrors CLIP’s approach to aligning image-text pairs, but instead aligns time-frequency representations of the same signal.

\clearpage

\subsubsection*{Temporal sequence}
\paragraph{TS-TCC}
We follow the same architecture implementation with the losses, i.e., contextual and temporal contrasting.
TS-TCC proposed applying two separate augmentations, one augmentation is weak (jitter-and-scale) and the other is strong (permutation-and-jitter). 
The authors also change the strength of the permutation window from dataset to dataset. 
In our experiments, we first used the original augmentations for each time series task, however, we observed performance decreases depending on the signal type.
We, therefore, applied the specific augmentations for each time series, where we observed a general performance improvement in other SSL techniques as well.

\paragraph{TS2Vec} 
TS2Vec~\cite{ts2vec} is a SSL method specifically designed for time series based on contrastive (instance and temporal wise) learning in a hierarchical way over augmented context views where the context is generated by applying timestamp masking and random cropping on the input time series.
Following the original framework, we use a dilated CNN architecture with a depth of 10 and hidden size of 64, which has a similar number of parameters with the architectures used by other SSL methods. 
The batch size is set to 8 and the number of epochs to 120, following the original manuscript.
Although we experimented with larger batch sizes, as SSL methods often benefit from them, we observed no performance gains.

\paragraph{TF-C}
The Time-Frequency Consistency (TF-C) method~\cite{tf_consistent} introduces a self-supervised learning framework for time series data by aligning representations from both time and frequency domains using the absolute Fourier transform.
TF-C employs specific augmentations in both domains, such as jittering in the time domain, and perturbations like adding or removing/decreasing frequency components in the Fourier domain, to create diverse views.
During inference, TF-C utilizes both the time-domain and frequency-domain encoders, combining their outputs to form the final representation, thereby integrating information from both domains for downstream tasks.
In contrast, our method avoids this by requiring only a single encoder at inference. We use the original implementation provided at \href{https://github.com/mims-harvard/TFC-pretraining}{github.com/mims-harvard/TFC-pretraining}.

\paragraph{SimMTM}
SimMTM~\cite{dong2023simmtm} presents a masked modeling framework tailored for temporal data. 
In time series, semantic information is heavily embedded in temporal variations, and random masking may disrupt critical patterns, making reconstruction unnecessarily difficult.
SimMTM mitigates this by treating masked modeling as a manifold learning problem.
Instead of reconstructing masked points directly from nearby unmasked values, it recovers them via weighted aggregation from multiple complementary masked sequences.
We use the same ResNet backbone for SimMTM to ensure fair comparison with other SSL methods. 
We also experimented with the original backbone proposed in the paper but observed no performance improvement.
For our evaluation, we follow the original manuscript's hyperparameter settings for the masking ratio and the number of positive (masked) series.
In addition, we explore the effects of training dynamics by running SimMTM with our higher batch size (1024) compared to the paper’s smaller batch sizes (128–256), and we test both short training epochs (40–50 as used in the original) and longer schedules, treating these as part of a hyperparameter search. 
We observed performance degradation when reducing the number of epochs, so we run all baseline methods for the same number of epochs to ensure a fair comparison.
We use the official implementation provided at \href{https://github.com/thuml/SimMTM}{github.com/thuml/SimMTM}.

\clearpage

\subsection{Implementation Details}
\label{appendix:Implementation_Details}
Here, we have provided the details of the architectures, and hyperparameters.
Primarily, we used the 1D ResNet~\cite{resnet1d} implementation in the supervised settings.
While some alternative deep learning models can perform better in a specific time series tasks such as the combination of convolutional and LSTM layers~\cite{CorNET, BeliefPPG}, we focused on residually connected convolutional architectures as backbones for representation learning.  

\subsubsection{Architectures}
Here, we present the details of architectures that are investigated for the performance of shift-invariant techniques.
Some details that are not given in the tables are as follows.
Batch normalization~\citep{batch} is applied after each convolutional block.
ReLU activation is employed following batch normalization, in line with~\cite{ResNet}.
We also applied a Dropout~\cite{Dropout} with 0.5 after each activation and before the convolutions.
Finally, a global average pooling is implemented before the linear layers.

\begin{table}[h]
    \centering
    \caption{Model architectures used in our experiments}
    \label{tab:resnet_unet}
    \begin{subtable}[t]{0.4\textwidth}
        \centering
        \caption{ResNet architecture for the main encoder}
        \label{appendix_resnet}
        \renewcommand{\arraystretch}{0.9}
        \begin{tabular}{llll}
            \toprule
            \# Blocks & Layer & Kernel & Output  \\
            \midrule
            1 & Input (C,T) & - & (C, T)  \\
            1 & Conv & (5, 1) & (64, T/2)  \\
            \midrule
            \multirow{2}{*}{8} & Conv & (5, 1) & (128, T/4) \\
            & Conv & (5, 1) & (128, T/4)  \\
            \midrule
            1 & Linear & - & (n\_classes,)  \\
            \midrule
            \multicolumn{4}{l}{\# Parameters for \textit{dataset} (C,F,T)} \\
            \multicolumn{3}{l}{\textit{IEEE SPC \& Dalia} (C=1, T=200)} & $\approx$210k \\
            \multicolumn{3}{l}{\textit{Chapman \& CPSC} (C=4, T=1000)} & $\approx$197k \\
            \multicolumn{2}{l}{\textit{Clemson} (C=1, T=240)} & & $\approx$200k \\
            \multicolumn{2}{l}{\textit{HHAR} (C=6, T=51)} & & $\approx$200k \\
            \multicolumn{2}{l}{\textit{USC} (C=6, T=100)} & & $\approx$200k \\
            \multicolumn{2}{l}{\textit{Sleep} (C=1, T=3000)} & & $\approx$210k \\
            \bottomrule
        \end{tabular}
    \end{subtable}
    \hfill
    \begin{subtable}[t]{0.45\textwidth}
        \centering
        \caption{Architecture for $\mathcal{W}(a,b)$}
        \label{tab:unet_2d_arch}
        \renewcommand{\arraystretch}{1}
        \begin{tabular}{lll}
            \toprule
            Layer & Input $\rightarrow$ Output & Description \\
            \midrule
            Input & $(C, F, T)$ & Input \\
            AvgPool & $\rightarrow (C, F/2, T/2)$ &  \\
            AvgPool2 & $\rightarrow (C, F/4, T/4)$ &  \\
            Conv1 & $\rightarrow (N, F, T)$ & Conv block \\
            Conv2 & $\rightarrow (2N, F/2, T/2)$ & \\
            Concat1 & $+\,\text{Pool1}$ & Merge \\
            Conv3 & $\rightarrow (3N, F/4, T/4)$ &  \\
            Concat2 & $+\,\text{Pool2}$ & Merge \\
            Conv4 & $\rightarrow (4N, F/8, T/8)$ & \\
            FC1 & $\rightarrow (1, 25)$ & Linear (time) \\
            FC2 & $\rightarrow (128)$ & Final output \\
            \midrule
            \multicolumn{3}{l}{\# Parameters for \textit{dataset} (C,F)} \\
            \multicolumn{2}{l}{\textit{IEEE SPC \& Dalia} (C=1, F=48)} & $\approx$500k \\
            \multicolumn{2}{l}{\textit{Chapman \& CPSC} (C=4, F=48)} & $\approx$500k \\
            \multicolumn{2}{l}{\textit{Clemson} (C=1, F=48)} &  $\approx$500k \\
            \multicolumn{2}{l}{\textit{HHAR} (C=6, F=48)} &  $\approx$550k \\
            \multicolumn{2}{l}{\textit{USC} (C=6, F=48)} &  $\approx$550k \\
            \multicolumn{2}{l}{\textit{Sleep} (C=1, F=48)} & $\approx$520k \\     
            \bottomrule
        \end{tabular}
    \end{subtable}
    
    \vspace{1mm} 

    \begin{subtable}[t]{0.45\textwidth}
        \centering
        \vspace{-1cm}
    \caption{Architecture for $\mathcal{F}(\mathrm{x})$}
    \label{tab:fourier_encoder}
    \renewcommand{\arraystretch}{1.0}
    \begin{tabular}{lll}
        \toprule
        Branch & Layer & Output \\
        \midrule
        \multirow{4}{*}{Amplitude} 
        & Conv1D & (16, T) \\
        & Residual  & (32, T/2) \\
        & Residual  & (64, T/4) \\
        & Linear  & (64) \\
        \midrule
        \multirow{4}{*}{Phase} 
        & Conv1D  & (16, T) \\
        & Residual  & (32, T/2) \\
        & Residual  & (64, T/4) \\
        & Linear & (64) \\
        \midrule
        Output & Concatenate & (128) \\
        \midrule
        \multicolumn{2}{l}{All datasets} & $\approx$55k \\     
        \bottomrule
    \end{tabular}
    \end{subtable}
    \hfill
    \begin{subtable}[t]{0.4\textwidth}
        \centering
        \caption{Architecture for mappers $\Phi^{t \rightarrow d}$}
        \label{tab:mapping}
        \renewcommand{\arraystretch}{1}
            \begin{tabular}{lll}
                \toprule
                Layer & \begin{tabular}[l]{@{}l@{}}Kernel\\Size\end{tabular} & \begin{tabular}[l]{@{}l@{}}Output\\Size\end{tabular} \\
                \midrule
                Input (1, L) & -- & (1, L) \\
                Conv1D & (3, 1), stride=2 & (64, L/2) \\
                ReLU & -- & (64, L/2) \\
                Conv Transpose & (3, 1), stride=2 & (1, L) \\
                \midrule
                \multicolumn{2}{l}{All datasets} & $\approx$500 \\
                \bottomrule
            \end{tabular}
    \end{subtable}
\end{table}

Our latent space mappers are lightweight, with each containing approximately 500 parameters, resulting in a total of only 1k additional parameters during inference.
This keeps the overall inference cost of our method low.
Moreover, even when baseline models are scaled up to match or exceed the total parameter count of  all encoders, they still fall short in performance, highlighting the effectiveness of our approach over brute-force model scaling.

\clearpage
\subsubsection{Augmentations}
We applied commonly used augmentations for each task, with details listed in Table~\ref{tab:appendix_augmentations}.
In each epoch, two random augmentations were applied per sample and instance discrimination is applied.

For the \textit{heart rate prediction} task, we used: \texttt{permutation}, \texttt{noise}, \texttt{scale}, and \texttt{shift}.

For \textit{activity recognition} and \textit{step detection} from inertial signals, we used: \texttt{permutation}, \texttt{noise}, \texttt{scale}, \texttt{shift}, and \texttt{rotation}.

For \textit{cardiovascular disease} and \textit{sleep stage} classification, we used: \texttt{resample}, \texttt{noise}, \texttt{scale}, \texttt{negate}, and \texttt{shift}.

These augmentation sets follow prior work~\cite{demirel2023chaos,CLOCS, KDD_paper, sivakumar2024emgqwerty}, ensuring task-relevant diversity.

\begin{table}[h]
\renewcommand{\arraystretch}{1.5}
\centering
\caption{Common time series augmentations }
\begin{adjustbox}{width=1\columnwidth,center}
\label{tab:appendix_augmentations}
\begin{tabular}{lll}
\hline
Domain & Augmentation & Details \\
\hline
\multirow{3}{*}{\textit{Time}} & Noise & Add Gaussian noise sampled from normal distribution, 
 $\mathcal{N}(0, 0.4)$  \\
 & Scale & Amplify channels by a random distortion sampled from normal distribution  $\mathcal{N}(2, 1.1)$  \\
 & Negate & Multiply the value of the signal by a factor of -1 \\
 & \multirow{2}{*}{Permute} & Split signals into no more than 5 segments, then permute the segments \\
 &                          & and combine them into the original shape \\
 & \multirow{2}{*}{Resample} & Interpolate the time-series to 3 times its original sampling rate \\
 &                           & and randomly down-sample to its initial dimensions \\
 & \multirow{2}{*}{Rotation} & Rotate the 3-axial (x, y, and z) readings of each IMU sensor by a random degree, which follows a \\
 &                           & uniform around a random axis in the 3D space. (Only applied for \textit{Activity Recognition})\\
 & Time Flip & Flip the time series in time for all channels, i.e., $\boldsymbol{\mathrm{x}}_{Aug}[n] = \boldsymbol{\mathrm{x}}[-n] $\\
 & Shift & Apply circular shift with a random amount\\
 & Random Zero Out & Randomly chose a section to zero out\\
 & Permutation + Noise & Combination of Permutation and Noise \\ 
 & Noise + Scale & Combination of Noise and Scaling\\ 
\midrule
\multirow{3}{*}{\textit{Frequency}}  & Highpass & Apply a highpass filter in the frequency domain to reserve high-frequency components\\ 
    & Lowpass & Apply a lowpass filter in the frequency domain to reserve low-frequency components \\  
    & Phase shift & Shift the phase of time-series data with a randomly generalized number\\
    & Noise in Frequency & Add Gaussian noise, sampled from normal distribution $\mathcal{N}(0, 0.5)$, to the frequency spectrum\\
\hline
\end{tabular}
\end{adjustbox}
\end{table}

Although we also experimented with frequency domain augmentations given in Table~\ref{tab:appendix_augmentations} for each task, we observed consistent performance degradation.
As a result, we excluded these frequency augmentations from all baseline comparisons.

\subsubsection{Transformations}
\label{appendix:Implementation_Details_transformations}

\paragraph{Fourier transformation}
For the Fourier transformation, we compute the FFT using the full length of each input signal without applying any task-specific padding or optimization.
Since the input signals are real-valued, we calculate only the positive (real) frequencies, leveraging the Hermitian symmetry of the spectrum.
We normalize the FFT using $\frac{1}{\sqrt{n}}$ by setting \texttt{norm='ortho'} in PyTorch’s~\cite{Torch} \texttt{torch.fft.rfft}, ensuring the transformation is orthonormal.

\paragraph{Wavelet transformation}
For the wavelet transformation, we use the continuous Morlet wavelet with 48 logarithmically spaced scales computed via \texttt{np.geomspace(1, 128, num=48)}, without any task-specific tuning or dataset-dependent adjustments.
This configuration is applied uniformly across all datasets to highlight the versatility and generality of our method.
We use PyWavelets~\cite{pywavelets} implementation.

\clearpage

\section{Computational Overhead of Techniques}
\label{sec:computational_overhead}
We analyze the computational overhead of SSL techniques designed for temporal data.
We break down the main operations of TS2VEC~\cite{ts2vec}, TS-TCC~\cite{tstcc}, TF-C~\cite{tf_consistent}, and simMTM~\cite{dong2023simmtm}, and provide empirical comparisons in Table~\ref{tab:time_comparison}.

\paragraph{TS2Vec}
TS2Vec samples overlapping subsegments for hierarchical contrastive loss.
For each sample, it randomly selects a crop length $l \in [2^{u+1}, T]$, defines a crop interval $[c_l, c_r]$ and an extended context interval $[e_l, e_r]$, and applies per-sample random offsets to generate two augmented views.
These views are passed through a shared encoder, producing two embeddings.
TS2Vec aligns their overlapping subsegments using a temporal contrastive loss and recursively applies pooling to compute multiple loss terms at increasingly coarser resolutions. 
This multiscale loss is recomputed for every pair at each layer depth, leading to quadratic scaling with the number of layers.
Moreover, because the crops are randomly sampled, TS2Vec forwards large sections of samples multiple times with different windows.
Despite using a shared encoder, this repeated window sampling and depth-wise contrastive loss accumulation make training slow, especially on long sequences.

\paragraph{TS-TCC}
In each batch, TS-TCC generates two augmented views using weak and strong augmentations, and processes both through a shared encoder to obtain temporal feature sequences.
The temporal contrasting module performs a cross-view prediction task: it uses an autoregressive model to summarize the past of one view into a context vector and predicts the future of the other view using linear projections. 
For each sample, this involves computing predictions for multiple future steps, comparing them against the target view, and accumulating a contrastive loss.
This adds sequential dependency to training, as the model must first encode the past then perform multiple forward passes to compare predicted and true representations.
While TS-TCC leverages efficient architectures like Transformers, the temporal prediction task and dual-view contrastive losses make it less parallelizable than methods that process views independently or avoid cross-timestep dependencies.

\paragraph{TF-C}
In each batch, TF-C computes time and frequency representations in parallel using separate encoders and applies multiple contrastive losses.
The input is first transformed via FFT to obtain frequency-domain features.
Two augmented views are generated: one in time via temporal augmentations (e.g., jittering, scaling), and one in frequency via spectral perturbations (adding/removing frequency components).
Both views are processed by their respective encoders and projectors.
TF-C then computes contrastive losses in three spaces: time-time, frequency-frequency, and time-frequency.
The time-frequency consistency loss includes four cross-domain pairings (e.g., original time vs. perturbed frequency) and is implemented using a triplet-style margin loss.
This results in eight forward passes per sample (two augmentations × two domains × two projections).

\paragraph{simMTM}
SimMTM introduces two coupled components: masked contrastive learning and masked reconstruction.
In each batch, a contrastive similarity matrix is computed over all samples and their masked variants, involving a full pairwise dot-product followed by calculating KL-divergence with soft targets.
The reconstruction further adds cost by aggregating weighted representations using the similarity matrix and applying a linear projection to reconstruct masked sequences.
While these components operate in parallel, the simultaneous use of large similarity matrices and reconstruction targets slows down training, especially with high masking rates or long windows.

Table~\ref{tab:time_comparison} reports the average execution time per epoch for each method, measured using an NVIDIA GeForce RTX 4090 GPU with 24GB of memory.
For each run, we timed only the core training operations using a unified timing function that accounts for both CPU and GPU execution.
The table presents the mean and standard deviation across the five runs.
For our method, the computation of wavelet and Fourier domain transformations for a batch of samples takes $\approx$ 2 seconds.
\begin{table}[b]
\vspace{-3mm}
    \centering
    \caption{Time taken (in seconds) for each SSL technique for a single epoch}
    \renewcommand{\arraystretch}{1.3}
    \vspace{0.7em} 
    \label{tab:time_comparison}
    \begin{tabular}{cccccc}
        \hline
        \textbf{Metric} & Ours & TS2Vec & TS-TCC & TF-C & simMTM \\ \hline
        Execution Time (sec) & \underline{18.67}\small$\pm$0.05 & 310\small$\pm$0.07 & \textbf{6.47}\small$\pm$0.09 & 21.61\small$\pm$0.02 & 68.14\small$\pm$0.99  \\ \hline
    \end{tabular}
\end{table}

\clearpage

A key advantage of our method is that the frequency-domain preprocessing (FFT and Gabor transforms) is performed only once and cached, rather than recomputed at every epoch.
In contrast, traditional temporal augmentations (e.g., permutation, scaling with noise) must be regenerated each epoch, introducing repeated overhead.

To quantify this difference, we report the runtime for a batch of 1024 samples (each being a univariate time series of length 200) measured on an NVIDIA RTX 4090 (Table~\ref{tab:preprocessing_cost}).
FFT and Gabor transforms take $0.0075$s and $3.59$s, respectively, but these costs are paid once and do not scale with the number of epochs.
By comparison, temporal augmentations scale linearly with training length: permutation requires $ 0.021 \times \text{epoch count} $ seconds and scale+noise requires $0.010 \times \text{epoch count}$ seconds.
At 500 epochs—common for SSL methods—our preprocessing yields about a $4\times$ speedup, and the advantage grows further for longer runs. 

\begin{table}[h]
    \centering
    \caption{Runtime of preprocessing steps (batch size 1024) measured on NVIDIA RTX 4090. Unlike augmentations, FFT and Gabor transforms are cached after one-time computation.}
    \vspace{0.7em} 
    \renewcommand{\arraystretch}{1.3}
    \label{tab:preprocessing_cost}
    \begin{tabular}{lcc}
        \hline
        \textbf{Transform} & \textbf{1 epoch (s)} & \textbf{500 epochs (s)} \\ \hline
        FFT & 0.0075 & 0.0075 \\
        Gabor & 3.59 & 3.59 \\
        Permutation & $0.021 \times \text{epoch\_count}$ & 10.5 \\
        Scale + Noise & $0.010 \times \text{epoch\_count}$ & 5.0 \\ \hline
    \end{tabular}
\end{table}

We note that our comparison focuses on the most commonly used temporal augmentations such as permutation and scaling with noise.
If we additionally implement and compare with the specialized time-series augmentation strategies~\cite{demirel2023chaos}, which often require auxiliary models or complex frequency-domain modifications, the computational gap becomes even larger.
In such cases, the advantage of our one-time preprocessing approach increases substantially, as these specialized augmentations add both runtime overhead and architectural complexity, whereas our method retains efficiency.

\clearpage

\section{Expanded Related Work}
\label{sec:expanded_related_work}
This section extends the related work discussed in the main paper by highlighting methods that aim to eliminate reliance on augmentations in self-supervised learning, as well as efforts to mitigate the representational biases introduced by common augmentations.

\paragraph{Representation invariance}
Representation learning methods have shown to be effective in several tasks.
Several works explain this success by relating the success of learned representations to invariancy caused by data augmentations~\cite{xiao2021what, aug_aware, yu2024selfsupervised, hamidieh2024views}.
However, such invariance could be harmful to downstream tasks~\cite{keller2023homomorphic, dangovski2022equivariant} if they rely on the characteristics of the data augmentations, e.g., location, amplitude-sensitive.
One proposed solution to this limitations involves constructing separate embedding spaces, each invariant to all but one augmentation~\cite{xiao2021what}.
While this approach allows for disentangling the effects of different augmentations, it is constrained by the number of predefined embedding spaces and comes with increased computational cost.
Another method introduces an augmentation-aware module that learns to predict the differences in augmentation parameters (e.g., cropping positions) between two randomly transformed samples~\cite{aug_aware}.

However, this technique has several drawbacks.
First, it requires explicit parameterization of each augmentation, which is particularly challenging for temporal data where augmentation semantics (e.g., shifts, warps) are less structured than in images.
Second, its effectiveness remains dependent on the choice and quality of augmentations used, reintroducing domain-specific tuning.

In contrast, our method leverages principled, isometric transformations that preserve the geometry of the data, without enforcing explicit invariance.
This allows the model to retain signal characteristics that may be important for downstream tasks and would otherwise be suppressed by augmentations~\cite{xiao2021what}.
For example, when shift invariance is beneficial in temporal tasks, our approach captures it naturally through domain projections such as the Fourier transform~\cite{demirel2025shifting}, without depending on handcrafted augmentations or specialized architectures~\cite{shift_invariant_old}.

\paragraph{Augmentation-free SSL}
Learning representations without relying on augmentations has been explored in domains where strong transformations risk distorting the data structure, such as graphs~\cite{augmentation_free_aaai, graph_augmentation_free} and time series~\cite{sui2024selfsupervised, temporal_freq_training}.
In graph learning, for instance, augmentation-free approaches generate alternative views by identifying nodes with similar local structures and global semantics~\cite{augmentation_free_aaai}.
However, these strategies are specialized for graph topologies and do not directly translate to temporal sequences.
More recently, authors in~\cite{sui2024selfsupervised} introduced random projections as a modality and application-agnostic alternative for self-supervised learning.
While this strategy provides a generic way to avoid manual augmentations, in practice, domain-aware contrastive methods equipped with carefully designed augmentations still tend to outperform projection-based approaches when reliable inductive biases about the data are available.

Our method goes a step further by eliminating the need for augmentations across diverse time-series applications, while still outperforming specialized augmentation strategies tailored to specific tasks, thereby highlighting its algorithmic superiority.

For time-series data, a recent approach introduced a temporal-frequency co-training model for semi-supervised learning, using time and Fourier domain transformations to generate pseudo-labels~\cite{temporal_freq_training}.
These methods generally overlook the non-stationary nature of temporal signals. 
Relying solely on Fourier transforms limits representation to global frequency patterns and may miss short-duration events.
In contrast, our method incorporates wavelet frames to capture local temporal variations, enabling finer resolution of transient signal components which are critical for representation learning.

Moreover, prior work often assumes that data should cluster similarly in time and frequency domains~\cite{temporal_freq_training}, implicitly treating the two latent spaces as geometrically aligned.
Our empirical results challenge this assumption, showing that latent representations across domains differ.
To address this, we introduce domain-specific latent space mappers that align and exploit the complementary structure of each space, enhancing representation learning.




\end{document}